\documentclass{article} 
\usepackage{iclr2026_conference,times}

\usepackage{amsmath,amsfonts,bm}

\def\eqref#1{equation~\ref{#1}}

\def\1{\bm{1}}

\DeclareMathAlphabet{\mathsfit}{\encodingdefault}{\sfdefault}{m}{sl}
\SetMathAlphabet{\mathsfit}{bold}{\encodingdefault}{\sfdefault}{bx}{n}

\usepackage{hyperref}
\usepackage{url}
\usepackage{booktabs} 
\usepackage{xcolor}         
\usepackage[table]{xcolor} 
\usepackage{graphicx}
\usepackage{multirow}

\usepackage{subcaption}
\usepackage{tabularx}
\usepackage{siunitx}
\usepackage{booktabs}
\usepackage{multirow}
\definecolor{mycolor_blue}{HTML}{E7EFFA}
\definecolor{mycolor_green}{HTML}{E6F8E0}
\definecolor{mycolor_gray}{HTML}{ECECEC}
\definecolor{mycolor_red}{HTML}{FFE6E6}
\definecolor{mycolor_yellow}{HTML}{FFFFCC}
\usepackage{wrapfig}
\usepackage{pifont}
\usepackage{tcolorbox}

\title{CodePlot-CoT: Mathematical Visual Reasoning by Thinking with Code-Driven Images}

\author{%
     \bf Chengqi Duan\textsuperscript{1}\thanks{Equal Contribution}
  ~~ \bf Kaiyue Sun\textsuperscript{1$\ast$}
  ~~ \bf Rongyao Fang\textsuperscript{3$\ast$}
  ~~ \bf Manyuan Zhang\textsuperscript{2}\thanks{Project Lead}
  ~~ \bf Yan Feng\textsuperscript{2}
  ~~ \bf Ying Luo\textsuperscript{2}\vspace{0.08cm}\\
    \bf Yufang Liu\textsuperscript{2}
  ~~ \bf Ke Wang\textsuperscript{3}
  ~~ \bf Peng Pei\textsuperscript{2}
  ~~ \bf Xunliang Cai\textsuperscript{2}
  ~~ \bf Hongsheng Li\textsuperscript{3}
  ~~ \bf Yi Ma\textsuperscript{1}
  ~~ \bf Xihui Liu\textsuperscript{1}\thanks{Corresponding Author}\vspace{0.3cm}
  \\
  \textsuperscript{1}HKU ~~~
  \textsuperscript{2}Meituan  ~~~
  \textsuperscript{3}CUHK ~~~
}

\iclrfinalcopy 
\begin{document}

\maketitle

\vspace{-9mm}
\begin{figure}[h]
    \makebox[\linewidth]{
        \includegraphics[width=1.0\linewidth]{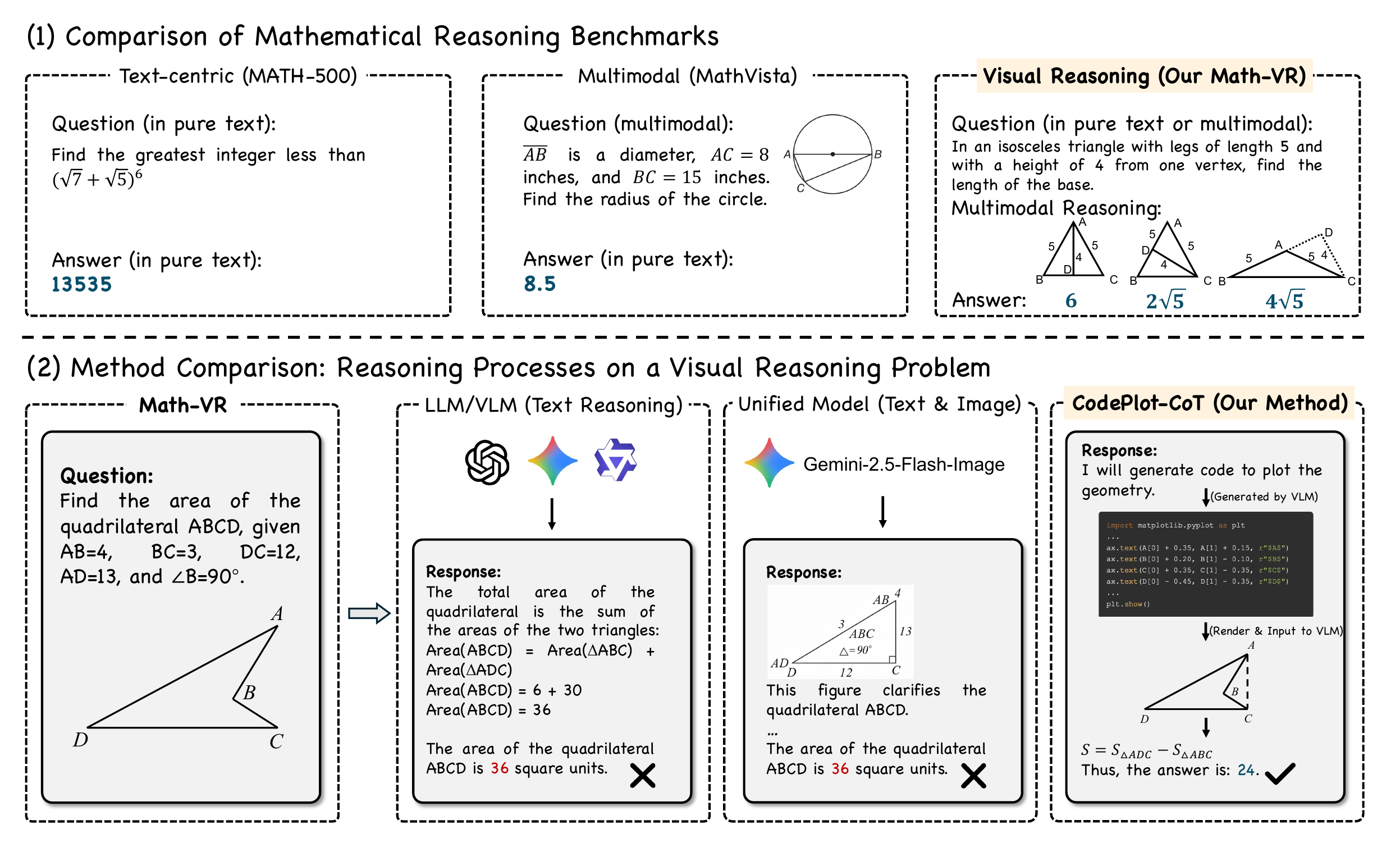}
    }
    \vspace{-6mm}
    \caption{\small A comparison of mathematical reasoning benchmarks and the methods on the visual reasoning problem. (1) illustrates that unlike existing benchmarks that rely on textual reasoning, Math-VR requires deep visual reasoning to resolve the math problems. (2) shows that on a visually ambiguous problem from Math-VR, both text-only and unified multimodal models fail. Our method, CodePlot-CoT, succeeds by programmatically generating the figure to uncover its true geometric properties, thus arriving at the correct solution.}
    \label{fig:teaser}
\vspace{-1mm}
\end{figure}

\begin{abstract}
\vspace{-2mm}
Recent advances in Large Language Models (LLMs) and Vision Language Models (VLMs) have shown significant progress in mathematical reasoning, yet they still face a critical bottleneck with problems requiring visual assistance, such as drawing auxiliary lines or plotting functions to solve the problems. Most LLMs and VLMs are constrained to text-only reasoning chains, while multimodal unified models that can generate interleaved text and images lack the necessary precision and controllability for such tasks. To address this, we propose \textbf{CodePlot-CoT}, a code-driven Chain-of-Thought paradigm for ``thinking with images'' in mathematics. Our approach leverages the VLM to generate text reasoning as well as executable plotting code, which is then rendered into images as ``visual thought'', to solve mathematical problems. To achieve this, we first construct \textbf{Math-VR}, the first large-scale, bilingual dataset and benchmark for Mathematics problems with Visual Reasoning, comprising 178K samples. Second, to create high-quality training data, we develop a state-of-the-art image-to-code converter specialized for parsing complex mathematical figures into codes. Finally, using these training data, we train the CodePlot-CoT model for solving mathematical problems. Experimental results show that our model achieves up to \textbf{21\%} increase over base model on our new benchmark, fully validating the efficacy of our proposed code-driven reasoning paradigm. Our work opens a new direction for multimodal mathematical reasoning and provides the community with the first large-scale dataset, comprehensive benchmark, and strong approach for such problems. To facilitate future research, we make our datasets, code, and pretrained models publicly
available at \url{https://github.com/HKU-MMLab/Math-VR-CodePlot-CoT}.
\end{abstract}

\begin{quote}
$~$\hfill {\em ``Algebra is but written geometry and geometry is but figured algebra.''}

$~$\hfill -- Sophie Germain
\end{quote}

\section{Introduction}
\vspace{-2mm}
Human cognition is inherently multimodal, leveraging visual graphs, diagrams, and sketches to facilitate complex reasoning. This is particularly evident in mathematics, where such visual aids—from drawing auxiliary lines in geometric proofs to plotting functions—are essential for rendering abstract relationships concrete and making the reasoning process more intuitive~\citep{hu2024visual}. While recent advances in Vision Language Models (VLMs) have shown strong performance in mathematical reasoning~\citep{gao2023g,shi2024math,zhang2024mavis}, they typically reply on text-only reasoning chains. This becomes a major limitation for problems that require visual reasoning, where humans would simply sketch diagrams or add auxiliary lines to facilitate reasoning. Such deficiency in multimodal reasoning leads to redundant and even incorrect text-only reasoning in mathematical problem solving (shown in Figure~\ref{fig:teaser}).



Recent efforts in general-domain visual understanding have explored the paradigm of Visual Chain-of-Thought (Visual CoT)~\citep{shao2024visual}, attempting to realize it by directly generating and manipulating images~\citep{li2025imagine,chen2025mint,li2025zebra}. However, this paradigm breaks down in the context of mathematics problems that demand high precision, where naive image generation is insufficient.  Even state-of-the-art unified models struggle to execute precise operations in mathematics, such as constructing auxiliary lines that satisfy strict geometric constraints.


The fundamental challenge in direct image generation and manipulation arises from the inherent difficulty in modeling the high-dimensional distribution of natural images, which contain complex textures and high-frequency details. However, mathematical visual aids differ significantly from general image generation tasks: the critical aspect is not the pixel-level details or textures, but rather the precise representation of key structured geometric properties such as shapes, lengths, positions, and angular relationships. This distinction suggests that the essential information in mathematical figures can be better captured by structured representations rather than pixel-level encodings. Therefore, we introduce programmatic code as the optimal representation for mathematical visual reasoning. As an inherently textual and structured representation format, code aligns seamlessly with language models~\citep{suris2023vipergpt,wang2025mathcoder,wang2023mathcoder}, enabling straightforward generation without introducing complex distributional modeling (\textit{e.g.}, diffusion models). 


In this work, we propose a new paradigm that enables VLMs to engage in visual reasoning through code generation. Instead of directly generating images with VLMs, which typically leads to degraded quality and precision in math plots, our approach guides the model to output executable plotting codes which are rendered into images as intermediate ``visual thoughts''. Once executed, the generated code produces images that can be input back into the VLM reasoning sequence.

Implementing this paradigm requires addressing two key challenges. First, there is a lack of structured dataset and benchmark for mathematical problems that demand visual reasoning, as existing works focus mainly on interpreting given figures rather than reasoning with visual images during problem solving~\citep{lu2023mathvista,zhang2024mathverse}. Therefore, we construct Math-VR, a large-scale bilingual dataset and benchmark comprising 173K training, 5K testing mathematical problems with visual reasoning solutions. We then benchmark existing SOTA models, thereby establishing strong baselines and highlighting the difficulty of this new task. Second, training models to output code as representations of visual thought requires a bidirectional mapping between code and images. We tackle this by developing MatplotCode, a high-fidelity image-to-code converter, which we leverage to construct code-driven CoT for training. The curated data then serves as the foundation for training CodePlot-CoT, a model specialized for code-driven visual reasoning. Our experiments show that it achieves a up to 21\% increase over base model, validating the efficacy of our approach.

The main contributions of our work can be summarized as follows:
\vspace{-2mm}
\begin{itemize}
    \item We propose a novel and efficien paradigm that enables VLMs to engage in visual reasoning through code generation.
    \vspace{-1mm}
    \item We construct \textbf{Math-VR}, the first large-scale, bilingual dataset and benchmark (178K samples) for Mathematical problems with Visual Reasoning.
    \vspace{-1mm}
    \item We develop \textbf{MatplotCode}, a state-of-the-art image-to-code converter for mathematical figures, and train \textbf{CodePlot-CoT} model, a specialized model that achieves up to a 21\% performance increase over strong baselines.
\end{itemize}
\section{Related Work}


\paragraph{VLMs in Mathematics Reasoning.} Reasoning greatly enhances model performance in general tasks.~\citep{wei2022chain,fang2025got,duan2025got} Recent progress on multimodal mathematical reasoning largely follows two lines: scaling math‑specific multimodal data and improving architectures for visual–text alignment. Representative data‑centric efforts include G‑LLaVA~\citep{gao2023g} with a geometry‑focused corpus (Geo170K), Math‑LLaVA~\citep{shi2024math} with the large‑scale MathV360K, and MAVIS~\citep{zhang2024mavis} which further optimizes math‑specific visual encoding and provides auto‑generated CoT‑rationales.  To evaluate these developments, benchmarks such as MathVista~\citep{lu2023mathvista}, MathVerse~\citep{zhang2024mathverse}, Math-Vision~\citep{wang2024measuring}, and MV-Math~\citep{wang2025mv} have emerged, each targeting different aspects of mathematical reasoning in visual contexts. Despite these advances, existing studies have mainly focused on understanding given visual inputs, rather than  incorporating visual information into the reasoning chain (plotting auxiliary lines or functions in solutions etc.).

\paragraph{``Thinking with image" Models.} To overcome the limitations of text-only reasoning, recent research focuses on ``Thinking with image," or Visual Chain-of-Thought (VCoT), where models actively retrieve or generate visual aids in reasoning process. Several  works~\citep{corbiere2025retrieval,jiang2025vlm,zhang2025cmmcot,GPT-o3,suris2023vipergpt,shao2024visual} implement a multimodal chain of thought where they retrieve and crop from the input image and interleave these visual snippets into the reasoning chain to provide more focused context for subsequent VQA steps. Other works~\citep{li2025zebra,li2025imagine,chern2025thinking,pan2025unlocking} focus on building unified models capable of generating interleaved text and image reasoning chain auto-regressively. These approaches discretize or embed images into visual tokens and train a single sequence model that sequencially outputs text and image during decoding. The generated images can serve as visual feedback for navigation tasks like mazes.



\paragraph{Visual Reasoning Models in Mathematics.}  Contemporary visual reasoning models for mathematics predominantly follow two paradigms: interleaved “thinking with image”~\citep{li2025zebra,chen2025mint,wang2025visuothink} and agent‑plus‑code tool use~\citep{hu2024visual,gao2023pal,zhou2023solve}. Interleaved approaches enable the model to sequentially add auxiliary lines and plot functions. However, the visual actions are weakly controllable, which hampers precise geometric constructions and limits the interpretability of intermediate reasoning steps. The agent‑plus‑code paradigm treats the model as a planner that create and manipulate input figure by predicting code snippets and call to external tools (Python, CAS/solver libraries, plotting utilities etc.). The executed outputs are again inputted to the model as visual feedback. In contrast to interleaved generation, tool-augmented agents provide precise and verifiable outputs by executing code, yet their performance largely depends on the reliability of the planner. Current planners are often zero-shot and not specifically trained for mathematical reasoning, which makes them susceptible to producing fragile or incorrect tool-use sequences.

\section{Math-VR: Dataset and benchmark for Math Visual Reasoning}

\subsection{Mathematical Problems require Visual Reasoning}
Previous mathematical benchmarks, such as Math-500~\citep{math_500}, AIME~\citep{AIME}, and HMMT~\citep{HMMT}, primarily evaluate models' textual reasoning abilities. More recent efforts like MathVista~\citep{lu2023mathvista} and MATH-Vision~\citep{wang2024measuring} introduce a multimodal setting by incorporating image-based questions. However, their focus remains largely on visual perception and extracting information from images, and the reasoning processes are still text-only reasoning without introducing visual thoughts. 

We argue that visual reasoning in mathematics should be an active process of ``reasoning with images'', which motivates us to propose a new benchmark, Math-VR. Figure~\ref{fig:teaser}(a) illustrates the distinction between existing benchmarks and ours. Previous benchmarks' reasoning can be performed entirely in text, whereas Math-VR necessitates multimodal reasoning with images. For example, the isosceles triangle problem shown in Figure~\ref{fig:teaser}(a) requires considering three possible scenarios.
Moving beyond naive visual perception, Math-VR demands solvers to conduct reasoning in both text and image domains, such as adding auxiliary lines, to assist with solving math problems. 


\begin{wrapfigure}{r}{0.5\textwidth}
\vspace{-30pt} 
\centering
\begin{tcolorbox}[colframe=black, colback=white!20, arc=5mm, boxrule=0.5mm, width=\linewidth]
\begin{tabular}{p{\linewidth}}

\textbf{Question:}

\vspace{0.5em}
As shown, $AB$ is a tangent to $\odot O$ at point $B$, and the extension of $AO$ meets $\odot O$ at point $C$.
If $\angle A = 45^\circ$ and $AB = 2$, then $AC$ equals (\quad)?

\includegraphics[width=0.35\linewidth]{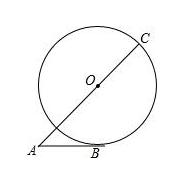}

\textbf{Analysis:}

\vspace{0.5em}
Connect $OB$, then $\triangle AOB$ is a right triangle.
Using trigonometry, we can find the length of $OA$, so $AC$ can be determined.

\includegraphics[width=0.35\linewidth]{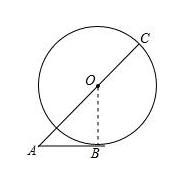}

Since $AB$ is a tangent to $\odot O$ at point $B$, Thus $OB \perp AB$.

In right triangle $OAB$: $OB = AB \cdot \tan A = 2 \times \tan 45^\circ = 2$; So $OA = \sqrt{OB^2 + AB^2} = \sqrt{2^2 + 2^2} = \sqrt{8} = 2\sqrt{2}$, So $AC = OA + OC = OA + OB = 2\sqrt{2} + 2$. 

Therefore, the answer is $2\sqrt{2} + 2$.

\end{tabular}
\vspace{-5pt}
\end{tcolorbox}
\vspace{-5pt}
\caption{Visualization of Math-VR sample.}
\label{fig:dataset_main}
\vspace{-20pt}
\end{wrapfigure}

\subsection{Dataset Construction}
\label{section3_2}
\textbf{Dataset Collection and Filtering.} We begin with collecting 900k secondary-school–level math problems with solutions from public websites, each containing at least one image in explanation (reasoning process) of the solution. Using Qwen2.5-VL-72B, we filter out irrelevant or purely textual images and retain only samples that require mathematical figures for reasoning. Furthermore, we employ GPT-4.1 to convert textual images into readable text and standardize each question into Markdown format. GPT-4.1 further conducts quality checks to discard incomplete or incoherent questions. 
This process results in \textbf{Math-VR dataset}, the first large-scale dataset targeted for visual mathematical reasoning, comprising 178,150 bilingual (English and Chinese) samples.

\textbf{Dataset Statistics.} In Math-VR dataset, each sample consists of a question, a reasoning process, and a final answer, with at least one image in the reasoning process. Math-VR encompasses a wide variety of visual reasoning tasks, with 29\% text-only and 71\% multimodal questions, spanning domains such as Geometry, Algebra, Calculus, and Statistics, where Geometry dominates (81\%), and is hierarchically categorized into subdomains and knowledge points (\textit{e.g.}, Triangle, Circle, Quadrilateral, Area, and Perimeter). More details about our dataset collection, categorization, and statistics are presented in Appendix.

\textbf{Visualization of Math-VR dataset.} Figure~\ref{fig:dataset_main} presents a visualization of Math-VR sample. Each sample in Math-VR contains both textual and visual reasoning. The visual figures in the analysis depict key geometric relations or intermediate steps, helping to clarify the reasoning process and demonstrate how image-based reasoning complements textual explanation. Additional examples are shown in Figure~\ref{fig:dataset_mm} and\ref{fig:dataset_text}  in Appendix.




\begin{figure}[t]
\vspace{-1cm}
\centering
 \begin{minipage}[t]{0.49\textwidth} 
 \centering
 \vspace{-29mm}
  \captionof{table}{\textbf{Key Statistics for Math-VR Benchmark.} We report statistics of our benchmark, including token lengths of questions and solutions, as well as the number and resolution of images.}
 \label{tab:statistics_benchmark}
\resizebox{0.7\linewidth}{!}{
\begin{tabular}{lc}
    \toprule
    \textbf{Statistics} & \textbf{Number}  \\
    \midrule
     \textbf{Question Length (text tokens)} \\
      \ - Minimum & 9 \\
      \ - Maximum & 602 \\
      \ - Average & 144.23 \\
     \midrule
     \textbf{Solution Length (text tokens)} \\
      \ - Minimum & 46 \\
      \ - Maximum & 2753 \\
      \ - Average & 591.14 \\
      \midrule
     \textbf{\# Images in Each Question} \\
      \ - Maximum number & 4 \\
      \ - Average number& 1.04 \\
      \ - Average resolution & 320x320 \\
      \midrule
     \textbf{\# Images in Each Solution} \\
      \ - Maximum number & 7 \\
      \ - Average number & 1.24 \\
      \ - Average resolution & 305x305 \\   

    \bottomrule
    \end{tabular}}

 \vfill
 \end{minipage} 
 \hfill
 \begin{minipage}[ht]{0.49\textwidth}
 \centering
 \vspace{1cm}
\includegraphics[width=0.8\linewidth]
{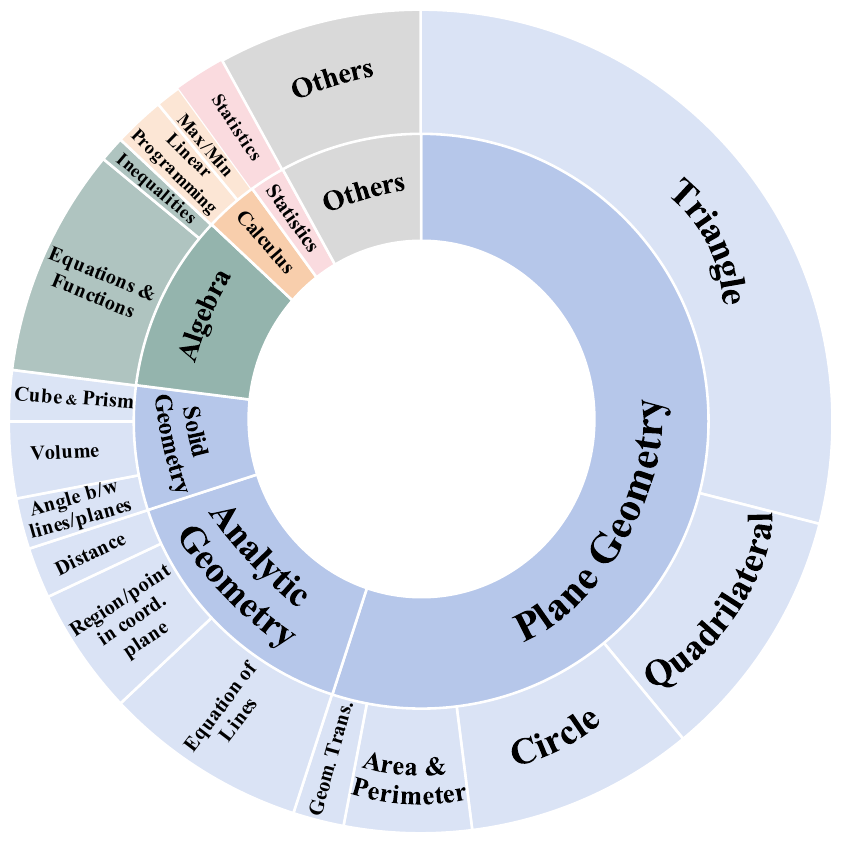}
 \caption{\textbf{Distribution of Knowledge Types in Math-VR Benchmark.} Geometry constitutes the majority of problems (77\%), with Algebra and Calculus comprising 13\%. } 
 \label{fig:benchmark_category}
 \end{minipage}
 \vfill
\end{figure}
\subsection{Benchmark and Evaluation}
\label{section3_3}
\textbf{Benchmark Construction.} To evaluate the visual mathematical reasoning capabilities of different models, we develop the Math-VR benchmark, which consists of 5k bilingual mathematical questions drawn from our dataset. 
The questions in Math-VR are selected through a careful pipeline designed to ensure
a deterministic and reliable evaluation. First, proof-based questions are excluded to avoid the difficulty and bias of assessing the logical validity. Most multiple-choice questions are excluded, since random guessing can yield correct answers by chance. 
From the remaining questions, a random pool of 3,000 samples was drawn from our dataset and manually reviewed to remove questions that require minimal or trivial visual reasoning.

\textbf{Benchmark Statistics and Distribution.} Our Math-VR benchmark is divided into two subsets: the Text subset, containing 2k text-only questions, and the Multimodal subset, comprising 3k questions that are demonstrated with both text and mathematical images.
Both subsets require reasoning or imagination in the visual domain to solve the questions. 
Table~\ref{tab:statistics_benchmark} and Figure~\ref{fig:benchmark_category} summarize the statistics and distribution of knowledge types of Math-VR benchmark, respectively. 

\begin{figure}[t]
    \makebox[\linewidth]{
        \includegraphics[width=1.0\linewidth]{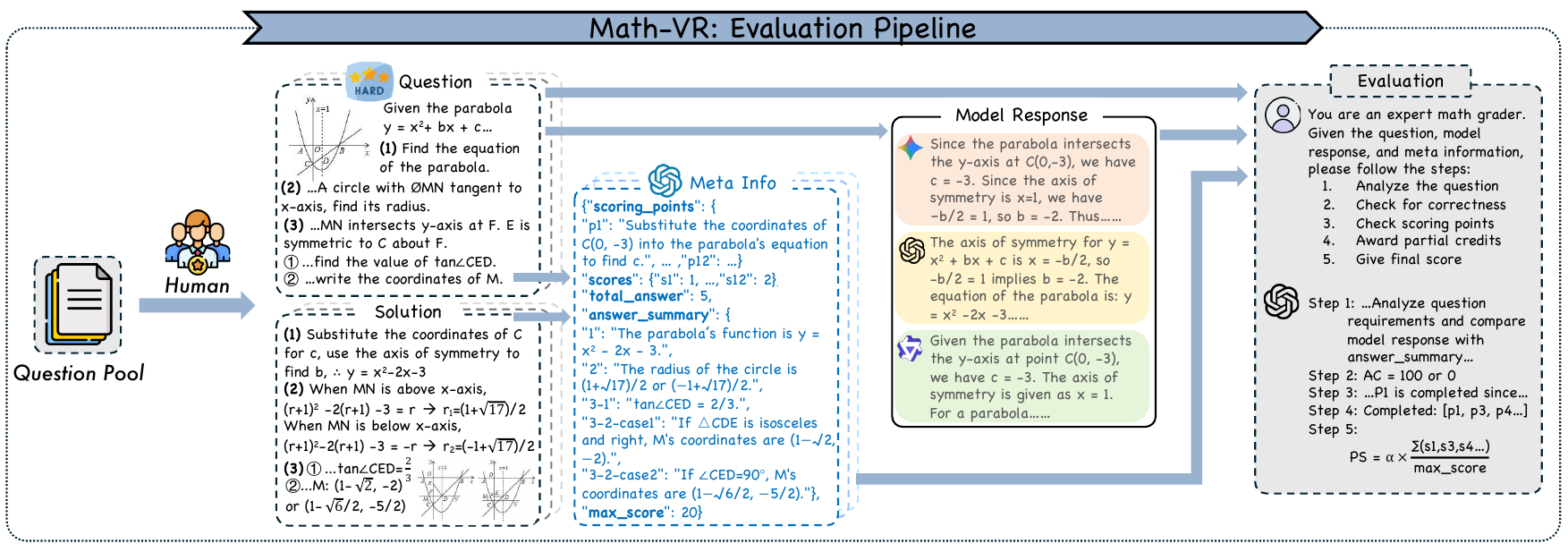}
    }
    \caption{\textbf{Math-VR Evaluation Pipeline}. 
    We design a VLM-based framework to comprehensively assess visual reasoning abilities of different models. The evaluation uses two metrics: Answer Correctness (AC), which gives a reliable binary judgment of the final answer, and Process Score (PS), which provides a fine-grained assessment of the solving process.}
    \label{fig:benchmark}
    \vspace{-10pt}
\end{figure}

\textbf{Evaluation Metrics.}
Figure~\ref{fig:benchmark} illustrates the evaluation pipeline of Math-VR, where GPT-4.1~\citep{GPT-4.1} is employed as the VLM evaluation tool. We design two metrics: Answer Correctness (AC) and Process Score (PS). 
Given the free-form nature of the answers (e.g., multiple numbers, ranges, or short text responses), we first use the VLM to analyze the ground-truth solution and generate a comprehensive summary of the final answer for each question.
Simultaneously, the VLM is prompted to identify ``scoring points'' within the solution. These scoring points refer to the critical steps required to solve the problem, such as applying theorems, making necessary deductions, and performing calculations. Each scoring point is assigned a value reflecting its difficulty (e.g., 1 or 2).
These extracted answers and scoring points serve as a reference to compare with the model-generated responses during evaluation.

(1) Answer Correctness (AC): 
To ensure consistent, reproducible, and objective evaluation results, this metric strictly checks whether the model-generated answer matches the ground-truth answer. If the answer is completely correct, it receives a score of 1 for AC; any error or omission results in a score of 0.

(2) Process Score (PS): When solving mathematical questions, even if the final answer is incorrect, the reasoning process may still be meaningful.
This metric awards partial credit if the model hits several scoring points in the reasoning process but fails to achieve the completely correct final answer.
If a final answer is completely correct (\textit{i.e.}, AC=100), then it automatically receives a PS of 100. Otherwise,
PS for a question $q$ is defined as follows: 
\begin{equation}
    \text{PS}(q) = 
        \alpha \times \dfrac{\sum_{j=1}^{m} v_j}{\sum_{i=1}^{n} v_i} \times 100, \text{ when the answer is not fully correct}
\label{equ2}
\end{equation}
where 
$\alpha$ represents a discount factor, we take $\alpha = 0.7$.
$n$ is the total number of scoring points for the question.
$m$ is the number of scoring points hit by the model answer.
$v_j$ is the point value of the $j$-th scoring point.

We have conducted a manual review of the VLM summarized information for each sample in our benchmark. For more details about the manual verification and the templates used for prompting GPT-4.1, please refer to the Appendix.

\begin{figure}[t]
    \makebox[\linewidth]{
        \includegraphics[width=1.0\linewidth]{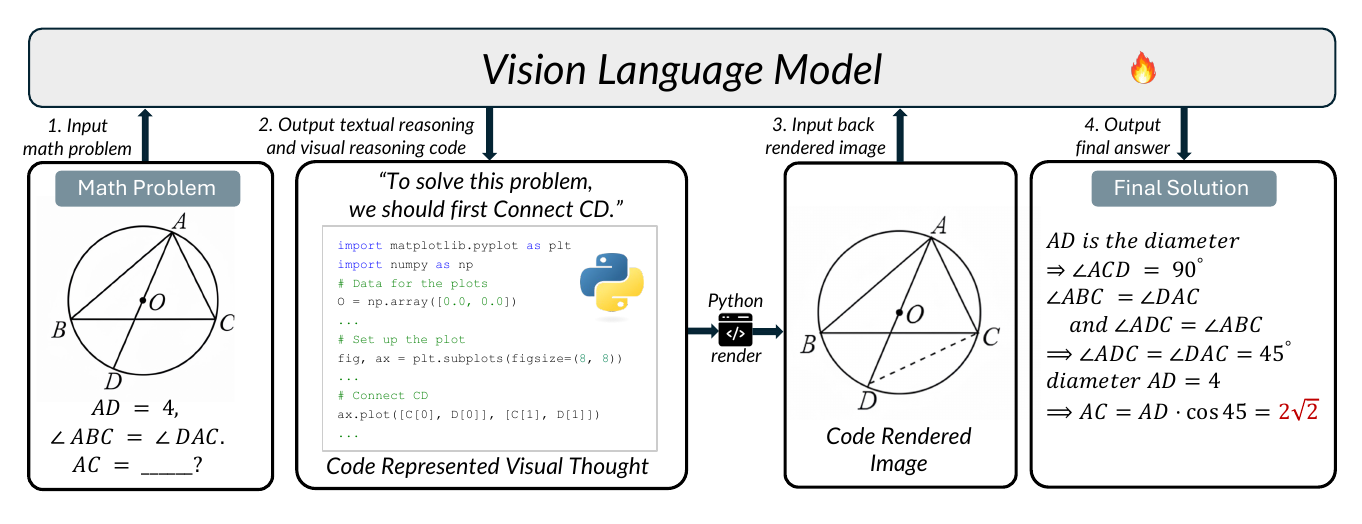}
    }
    \caption{\textbf{Illustration of the CodePlot-CoT paradigm for mathematical visual reasoning}. The model interleaves natural language reasoning with code-based visual reasoning. At points in the solution that require visual support, the model generates a sequence of executable plotting code, which is rendered via Python into precise figures and input back into the model. This allows the model to ``see'' its own visual thought and leverage it for subsequent reasoning, ultimately leading to a more accurate final solution. CodePlot-CoT supports interleave text and multi-image reasoning.}
    \label{fig:main_figure}
\end{figure}
\section{CodePlot-CoT Paradigm: Code-driven CoT for Mathematics Visual Reasoning}

\subsection{Paradigm Overview}


Existing VLMs remain constrained when visual aids are required. Most models still rely primarily on text-only chain-of-thought reasoning, which often fails to capture complex visual elements and geometric properties. To address this, recent efforts introduce Visual CoT by directly generating or manipulating images. However, it is difficult for current image generation models to satisfy strict geometric constraints in the mathematical context. These shortcomings constitute fundamental obstacles to accurate visual reasoning in mathematics. 

To overcome these limitations, we propose CodePlot-CoT, which represents ``visual thoughts'' as executable plotting code instead of pixel-encoded images. Our key insight is to replace pixel-level visual generation and manipulation with a language modeling problem: instead of ``drawing'' in visual space, the VLM can ``write'' code in the text modality where it is inherently proficient. Mathematical figures do not rely on pixel-level fidelity or texture details. Instead, what matters are the structural attributes such as geometric shapes, spatial positions, and angular relations, making executable plotting code a perfect fit for representing such structured geometric information. 
By releasing the burden of pixel-level distribution modeling, the model can concentrate on the precise geometric information, making it easier to reason with mathematical figures.

Figure~\ref{fig:main_figure} presents the CodePlot-CoT paradigm, where text reasoning is interleaved with code-based visual reasoning. The model first generates a reasoning chain in natural language. When visual reasoning step is necessary (such as constructing an auxiliary line), it generates a block of plotting code that represents the required visual information. This code is executed to render an image, which is then input back to the model as the visual reasoning. Such visual thought allows the model to ground its subsequent reasoning in precise, self-generated visual evidence. This fundamentally changes the nature of the model's reasoning, moving it beyond text-only reasoning to a multimodal reasoning chain where thoughts and hypotheses are proposed, tested, and refined in both visual and linguistic domains. 

\subsection{Code-Driven CoT Curation with Image-to-Code Converter}

A core prerequisite for training CodePlot-CoT is the high-quality data that integrates images, plotting code, and reasoning chains as presented in Figure~\ref{fig:main_figure}. Such data allows the model to learn how visual thoughts can be faithfully represented as executable code. However, existing mathematical resources rarely provide paired code annotations for visual reasoning, making it difficult to obtain the structured supervision required for code-driven CoT. This motivates the need for a reliable image-to-code mapping that can convert mathematical figures to plotting code. Yet, no fine-grained converter specialized for this domain is publicly available. Even large commercial models (e.g., Gemini-2.5-Pro, GPT-5) are unreliable for zero-shot image-to-code conversion on complex mathematical figures, limiting practical effectiveness. To overcome this bottleneck, we develop \textbf{MatplotCode}, a state-of-the-art converter tailored for mathematical figures, which enables scalable creation of code–image pairs and supports the curation of supervised fine-tuning data for CodePlot-CoT.


We leverage the ImgCode-8.6M dataset from MathCoder-VL~\citep{wang2025mathcoder} as the foundation of our experiment. We begin by filtering out images that are not representative of standard mathematical figures, thereby curating a high-quality subset focused on geometry diagrams and function plots. All code representations in this dataset are in Python. On this curated data, we train MatplotCode, which demonstrates superior generalization and conversion fidelity. To further enhance the quality of our supervised fine-tuning data, we use MatplotCode to generate multiple Python code representations for each source image and employ GPT-4.1 to select the optimal representation.


\subsection{Training Details}
\label{section_4_3}
 We leverage Qwen2.5VL-32B as the base model for both MatPlotCode and CodePlot-CoT model. MatPlotCode goes through a two-stage training process: we first align the visual components by training only the vision encoder (ViT) and the MLP projector for one epoch, and then perform full-parameter fine-tuning for two additional epochs. For the CodePlot-CoT model, we initialize its weights from vision-aligned MatPlotCode after Stage 1. We then fully finetune this model on our curated SFT dataset for 5000 steps. During MatPlotCode training, we assign loss to the textual reasoning chain and generated code while no loss is applied to the rendered images in the sequence. More details on our training are presented in the Appendix.

\section{Experiments}
\subsection{Human Correlation Analysis}
To further validate the reliability of our automatic evaluation pipeline, we conduct a human correlation study. We invite 15 senior undergraduate students majoring in STEM disciplines as our human experts. We then randomly sample 1000 questions from our benchmark and generate a total of 3000 answers from GPT4.1, Gemini-2.5-pro and Claude-Sonnet-4. Each participant is assigned 200 answers and is asked to (i) judge the Answer Correctness (AC) in a correct/incorrect manner and (ii) assign a Process Score (PS) based on the same scoring point in our benchmark. Both scores show strong consistency between human annotations and GPT-4.1 evaluations. 
\begin{itemize}
    \item \textbf{Answer Correctness (AC):} We report Cohen’s $\kappa = 0.75$ and MCC $= 0.75$. 
        For binary judgments, beyond simple accuracy, these measures account for chance agreement and balance between positive/negative classes. 
    \item \textbf{Process Score (PS):} We report Pearson $r = 0.72$ and Spearman $\rho = 0.70$. 
    Pearson captures the linear correlation between human and GPT-4.1 scores, 
    while Spearman focuses on rank-order consistency. 
\end{itemize}

\subsection{Benchmarking Existing Models and CodePlot-CoT on Math-VR}
For a comprehensive evaluation, we compare our approach against a suite of state-of-the-art LLMs, VLMs, or UMs, including both open-source and closed-source ones on the 2500 English questions in our benchmark, as shown in Table~\ref{tab:main_result}. Among closed-source and large open-source models, Gemini-2.5-Pro stands out, achieving the highest overall scores (PS = 80.8, AC = 64.7) on the Math-VR. We observe that ``thinking'' models which benefit from better-structured textual chains performs better on our benchmark. Among no-thinking models, Nano Banana achieves the highest score, primarily due to its stronger visual reasoning capabilities rather than purely text-based chains. Nevertheless, there remains substantial space for improvement in Answer Correctness across closed-source models.
Even the strongest performer, Gemini-2.5-Pro, still fails on approximately one-third of the benchmark problems. This gap highlights that current advances in chain length and scale alone are insufficient, and points toward future research directions in developing more faithful and controllable visual reasoning mechanisms.

The limitations are more significant in open-source models under 100B, most of which remain constrained to pure text reasoning and consequently exhibit low answer correctness on Math-VR benchmark (e.g., Gemma, Qwen2.5-VL-72B). Attempts to ``think with images'' by directly generating pixels (e.g., Bagel-Zebra-CoT) provide only modest gains due to poor controllability and low geometric precision due to model scale. In contrast, our code-driven approach yields substantial improvements: CodePlot-CoT surpasses its 32B base VLM by up to 21\% and largely outperforms the Qwen2.5-VL-72B across all metrics, demonstrating that structured, verifiable visual reasoning is more decisive than model size or longer textual chains. CodePlot-CoT generated examples are shown in Figure~\ref{fig:ours_mm} and \ref{fig:ours_text} in Appendix.

\begin{table}[htb]
\caption{
\textbf{Math-VR evaluation results.}
This table compares the model performances on our benchmark. The second column specifies the model size by its parameter count. For models that support an internal reasoning process, \checkmark in the ``Thinking'' column indicate this mode is turned on during evaluation. 
Model Types: VLM: Vision Language Model, LLM: Large Language Model, UM: Unified Model. 
Metrics: PS: Process Score, AC: Answer Correctness. 
\colorbox{mycolor_blue}{Blue} and \colorbox{mycolor_yellow}{Yellow} highlight the top score in each model group. \textbf{Bold} signifies the highest score across all models. 
}\label{tab:main_result}
\centering
\renewcommand{\arraystretch}{1.1} 
\resizebox{\textwidth}{!}{%
\begin{tabular}{l|c|c|c|cc|cc|cc}


\toprule
\multirow{2}{*}{\textbf{Model}} &\multirow{2}{*}{\textbf{\# Params}} &\multirow{2}{*}
{\textbf{Type}} & \multirow{2}{*}{\textbf{Think}} & \multicolumn{2}{|c|}{\textbf{Text}} & \multicolumn{2}{|c|}{\textbf{Multimodal}} & \multicolumn{2}{|c}{\textbf{Overall}} \\
\cmidrule{5-10}
& & & & PS & AC & PS & AC & PS & AC \\
\midrule
\multicolumn{10}{c}{\textit{Closed-source Models and Open-source Models over 100B}}\\ \midrule
\textbf{GPT-4o(\citeyear{GPT-4o})} & - & VLM & $\times$ & 34.6 & 5.7 & 27.6 & 3.4 & 30.4 & 4.3 \\
\textbf{GPT-4.1-nano(\citeyear{GPT-4.1})}        & - & VLM & $\times$ & 45.9 & 13.1 & 33.6 & 6.4 & 38.5 & 9.1 \\
\textbf{GPT-4.1-mini(\citeyear{GPT-4.1})}        & - & VLM & $\times$ & 62.0 & 33.3 & 58.6 & 33.3 & 60.0 & 33.3 \\
\textbf{GPT-4.1(\citeyear{GPT-4.1})}             & - & VLM & $\times$ & 56.5 & 26.6 & 52.2 & 25.6 & 53.9 & 26.0 \\

\textbf{GPT-o3(\citeyear{GPT-o3})} & - & VLM & $\checkmark$ & 72.9 & 52.9 & 78.6 & 63.7 & 76.4 & 59.3 \\
\textbf{Gemini-2.0-Flash(\citeyear{Gemini-2.0})}    & - & VLM & $\times$ & 56.1 & 24.1 & 47.0 & 18.3 & 50.7 & 20.6 \\
\textbf{Gemini-2.5-Flash(\citeyear{Gemini-2.5-flash})} & - & VLM & $\times$ & 70.9 & 44.6 & 75.5 & 57.5 & 73.7 & 52.3 \\
\textbf{Gemini-2.5-Flash(\citeyear{Gemini-2.5-flash})} & - & VLM & $\checkmark$ & 77.5 & 57.0 & 79.0 & 62.9 & 78.4 & 60.5 \\
\textbf{Gemini-2.5-Pro(\citeyear{Gemini-2.5-pro})}      & - & VLM & $\checkmark$ & \colorbox{mycolor_blue}{\textbf{77.9}} & \colorbox{mycolor_blue}{\textbf{58.7}} & \colorbox{mycolor_blue}{\textbf{82.8}} & \colorbox{mycolor_blue}{\textbf{68.7}} & \colorbox{mycolor_blue}{\textbf{80.8}} & \colorbox{mycolor_blue}{\textbf{64.7}} \\
\textbf{Nano Banana(\citeyear{nano-banana})} & - & UM & $\times$ & 72.3 & 49.1 & 74.7 & 56.3 & 73.8 & 53.4  \\
\textbf{Seed-1.6-Thinking(\citeyear{Seed1.6})}   & - & VLM & $\checkmark$ & 73.0 & 53.0 & 76.6 & 62.0 & 75.2 & 58.4 \\
\textbf{Claude-Sonnet-4(\citeyear{claude-sonnet-4})}     & - & VLM & $\times$ & 60.9 & 31.5 & 53.4 & 25.8 & 56.4 & 28.1 \\
\textbf{GLM-4.5V(\citeyear{glm})}            & 108B & VLM & $\checkmark$ & 70.5 & 48.0 & 69.1 & 50.6 & 69.7 & 49.6 \\
\textbf{Deepseek-R1\citeyear{guo2025deepseek}}  & 671B & LLM & $\checkmark$ & 69.9 & 49.5 & - & - & - & - \\
\midrule
\multicolumn{10}{c}{\textit{Open-source Models under 100B}}\\ \midrule

\textbf{Bagel(\citeyear{deng2025emerging})} & 7B & UM & $\times$ & 32.9 & 8.5 & 24.0 & 7.0 & 27.6 & 7.6 \\
\textbf{Bagel-Zebra-CoT(\citeyear{li2025zebra})} & 7B & UM & $\times$ & 41.5 & 13.9 & 29.1 & 7.6 & 34.1 & 10.1 \\
\textbf{Keye-VL-1.5(\citeyear{keye})} & 8B & VLM & $\times$ & 44.4 & 20.2 & 34.0 & 15.4 & 38.2 & 17.3  \\
\textbf{InternVL-3.5-8B(\citeyear{wang2025internvl3})} & 8B & VLM & $\times$ & 35.6 & 9.2 & 28.6 & 7.0 & 31.4 & 7.9       \\
\textbf{Gemma3(\citeyear{team2025gemma})} & 27B & VLM & $\times$ & 50.8 & 19.2 & 40.8 & 14.1 & 44.8 & 16.1 \\
\textbf{Qwen-2.5-VL-3B(\citeyear{bai2025qwen2})}  & 3B & VLM & $\times$ & 33.4 & 7.9 & 23.6 & 3.6 & 27.5 & 5.3 \\
\textbf{Qwen-2.5-VL-7B(\citeyear{bai2025qwen2})}  & 7B & VLM & $\times$ & 18.0 & 4.5 & 11.0 & 2.0 & 13.8 & 3.0 \\
\textbf{Qwen-2.5-VL-32B(\citeyear{bai2025qwen2})} & 32B & VLM & $\times$ & 36.9 & 10.6 & 31.5 & 9.6 & 33.7 & 10.0 \\
\textbf{Qwen-2.5-VL-72B(\citeyear{bai2025qwen2})} & 72B & VLM & $\times$ & 44.6 & 15.3 & 38.2 & 12.7 & 40.8 & 13.7 \\
\midrule
\textbf{CodePlot-CoT} & 32B & VLM & $\times$ & \colorbox{mycolor_yellow}{53.8} & \colorbox{mycolor_yellow}{31.6} & \colorbox{mycolor_yellow}{42.4} & \colorbox{mycolor_yellow}{15.8} & \colorbox{mycolor_yellow}{47.0} & \colorbox{mycolor_yellow}{22.1} \\
$\Delta$ Over Base Model & & &  & \textbf{+16.9} & \textbf{+21.0} & \textbf{+10.9} & \textbf{+6.2} & \textbf{+13.3} & \textbf{+12.1} \\
\bottomrule
\end{tabular}}
\vspace{-0.2cm}
\end{table}


\subsection{Image-Code Converter Evaluation}
We evaluate MatplotCode against FigCodifier-8B, GPT-o3 and Gemini-2.5-pro (thinking budget maximum). We randomly sample 1000 images from our dataset and task these models to convert it to matplotlib code. We assess two aspects: (i) Execution Success Rate, i.e., the probability that the generated code runs without errors.
(ii) Reconstruction Fidelity, judged by GPT-4.1 via a standardized prompt to decide which reconstruction is most similar to the original image.
The full evaluation prompt is provided in the Appendix.

Both MatplotCode and the FigCodifier-8B achieve a 100\% execution success rate.  In contrast, GPT-o3 reach 79.6\%, while Gemini-2.5-Pro achieve 86.2\%, indicating a higher likelihood of producing invalid or incomplete code. For reconstruction fidelity, judged by GPT-4.1, MatplotCode is preferred in 554 out of 1,000 cases, compared to 190 for FigCodifier-8B, 49 for GPT-o3, and 207 for Gemini-2.5-Pro.
These results demonstrate that our converter not only guarantees reliable code prediction, but also produces reconstructions that are consistently closer to the original figures. Importantly, the failures of both closed-source large models and open-source expert models further underscore the necessity of developing a new converter tailored to this task.

\subsection{Analysis on Inference Cost}
In this section, we analyze the inference cost of our code-driven visual reasoning paradigm. On the 2,500 test problems, our model generates an average of 820.9 tokens per image and 3,416 rendered images in total, or 1.37 images per problem in average. From the perspective of token usage, this cost is lower than many autoregressive “thinking with image” models, which typically consume 1,024 or even up to 4,096 tokens per image. Moreover, each image can be rendered locally in less than one second, making it negligible when evaluating overall computational efficiency. We further measure the total number of output tokens. On average, our model produces 567.2 text tokens, and together with the code tokens for rendered images, the overall output is 1,691.8 tokens per problem. In comparison, Qwen2.5-VL-32B generates substantially more content, averaging 3,847.3 tokens. This significant reduction in output length highlights the efficiency of our code-driven visual reasoning paradigm on Math-VR.

\subsection{Ablation Studies}
To better understand the contribution of each design choice in our framework, we conduct two sets of ablation experiments.

\paragraph{Text-only vs. Code-driven Visual Reasoning.}
We first compare our code-driven paradigm against text-only reasoning using Qwen-2.5VL-3B as the base model. 
To establish a text-only finetuned baseline, we remove all images from the solutions in our dataset and fine-tune the model exclusively on textual reasoning, resulting in Qwen-2.5VL-3B-Text-Tune. 
This model shows only marginal improvement over the vanilla baseline because it remains fundamentally constrained by its inability to incorporate visual information. 
In contrast, our CodePlot-CoT-3B achieves substantial gains, clearly demonstrating the advantage of introducing executable code for visual reasoning.

\paragraph{Code-driven vs. Direct Image Generation.}
We further investigate whether code-based visual reasoning is more effective than direct image generation in reasoning. 
Since Qwen2.5-VL does not support image generation, we perform this comparison on the unified model Bagel. 
The Bagel-Thinking-with-image is fine-tuned to directly produce interleaved text–image outputs on our dataset. 
Although this approach provides some improvements, it underperforms our CodePlot-CoT-Bagel, which leverages structured executable code. 
These results validate that code-driven reasoning offers a more precise and controllable representation of visual thoughts than direct pixel-level generation.


\begin{table*}[t]
  \centering
  \begin{minipage}[t]{0.48\textwidth}
    \vspace{0pt} 
    \centering
    \caption{Comparing our code-driven paradigm against text-only reasoning. Qwen-2.5VL-3B-Text-Tune is fine-tuned on text-only CoT, while CodePlot-CoT-3B is trained with our paradigm. The results demonstrate the significant performance gain from enabling code-based visual reasoning.}
    \resizebox{\textwidth}{!}{
    \begin{tabular}{c|cc|cc|cc}
    \toprule
    \multirow{2}{*}{\textbf{Ablation Setting}} & 
    \multicolumn{2}{c|}{\textbf{Text}} & 
    \multicolumn{2}{c|}{\textbf{Multimodal}} & 
    \multicolumn{2}{c}{\textbf{Overall}} \\
    \cmidrule(lr){2-7}
    & PS & AC & PS & AC & PS & AC \\
    \midrule
    Qwen-2.5VL-3B & 33.4 & 7.9 & 23.6 & 3.6 & 27.5 & 5.3\\
    Qwen-2.5VL-3B-Text-Tune & 34.3 & 12.7 & 27.4 & 5.8 & 30.1 & 8.4 \\
    CodePlot-CoT-3B & \textbf{35.5} & \textbf{13.6} & \textbf{29.3} & \textbf{7.4} & \textbf{31.8} & \textbf{9.9} \\
    \bottomrule
    \end{tabular}}
    \label{abla_qwen}
  \end{minipage}
  \hfill
  \begin{minipage}[t]{0.48\textwidth}
    \vspace{0pt} 
    \centering
    \caption{Comparing our code-driven paradigm against direct image generation VCoT. Bagel-Thinking-with-image is fine-tuned to generate direct image outputs in reasoning steps, while CodePlot-CoT-Bagel uses our code-generation paradigm. This validates the efficacy of our paradigm.}
    \resizebox{0.98\textwidth}{!}{
    \begin{tabular}{c|cc|cc|cc}
    \toprule
    \multirow{2}{*}{\textbf{Ablation Setting}} & 
    \multicolumn{2}{c|}{\textbf{Text}} & 
    \multicolumn{2}{c|}{\textbf{Multimodal}} & 
    \multicolumn{2}{c}{\textbf{Overall}} \\
    \cmidrule(lr){2-7}
    & PS & AC & PS & AC & PS & AC \\
    \midrule
    Bagel & 32.9 & 8.5 & 24.0 & 7.0 & 27.6 & 7.6 \\
    Bagel-Thinking-with-image & 40.3 & 10.1 & 28.4 & 8.3 & 33.2 & 9.0 \\
    CodePlot-CoT-Bagel & \textbf{43.1} & \textbf{11.9} & \textbf{31.1} & \textbf{10.2} & \textbf{35.9} & \textbf{10.9} \\
    \bottomrule
    \end{tabular}}
    \label{abla_bagel}
  \end{minipage}
\end{table*}
\section{Conclusion}
\vspace{-10pt}
In this work, we introduce CodePlot-CoT, a code-driven chain-of-thought paradigm that enables VLMs to “think with images” in mathematical reasoning. By representing visual reasoning as executable snippets of plotting code, our approach circumvents the limitations of pixel-level image generation, achieving precise and controllable visual thought. To achieve this paradigm, we construct Math-VR, the first large-scale bilingual dataset benchmark for mathematical visual reasoning, and develope MatplotCode, a high-fidelity image-to-code converter. Extensive experiments demonstrate that our model consistently outperforms baseline models with improvements of up to 21\%. We have used GPT-5 to help refine grammar in this paper.

\section{Ethics and Reproducibility Statement}
All data used in this study are collected from publicly available websites, ensuring that no private or sensitive information is involved in the dataset construction. The details on dataset, benchmark construction and evaluation are presented in Section ~\ref{section3_2} and Section ~\ref{section3_3}. Training and implementation details are described in Section ~\ref{section_4_3}. Additional information, including evaluation templates, manual verification processes, and further dataset construction details, is provided in the Appendix. These resources are made available to facilitate transparent assessment and to support reproducibility of our results.

\section{Acknowledgments}
We would like to extend our special thanks to the following members for their participation in the manual verification of the benchmark: Manyuan Zhang and Yan Feng from Meituan, Kunming Luo and Yexin Liu from The Hong Kong University of Science and Technology, Zihao Pan from Sun Yat-sen University, Dian Zheng, Rongyao Fang, Kaituo Feng, and Yilei Jiang from CUHK, Zhangquan Chen from Tsinghua University, Yimeng Jia from Peking University, Hongyu Li from Beihang University and Zhekai Chen and Chengqi Duan and 
Kaiyue Sun from The University of Hong Kong.

\bibliography{iclr2026_conference}

\begin{thebibliography}{45}
\providecommand{\natexlab}[1]{#1}
\providecommand{\url}[1]{\texttt{#1}}
\expandafter\ifx\csname urlstyle\endcsname\relax
  \providecommand{\doi}[1]{doi: #1}\else
  \providecommand{\doi}{doi: \begingroup \urlstyle{rm}\Url}\fi

\bibitem[Anthropic(2025)]{claude-sonnet-4}
Anthropic.
\newblock Introducing claude 4.
\newblock \url{https://www.anthropic.com/news/claude-4}, 2025.

\bibitem[Bai et~al.(2025)Bai, Chen, Liu, Wang, Ge, Song, Dang, Wang, Wang, Tang, et~al.]{bai2025qwen2}
Shuai Bai, Keqin Chen, Xuejing Liu, Jialin Wang, Wenbin Ge, Sibo Song, Kai Dang, Peng Wang, Shijie Wang, Jun Tang, et~al.
\newblock Qwen2. 5-vl technical report.
\newblock \emph{arXiv preprint arXiv:2502.13923}, 2025.

\bibitem[Chen et~al.(2025)Chen, Zhang, Jiang, Zhou, Yan, Lin, and Li]{chen2025mint}
Xinyan Chen, Renrui Zhang, Dongzhi Jiang, Aojun Zhou, Shilin Yan, Weifeng Lin, and Hongsheng Li.
\newblock Mint-cot: Enabling interleaved visual tokens in mathematical chain-of-thought reasoning.
\newblock \emph{arXiv preprint arXiv:2506.05331}, 2025.

\bibitem[Chern et~al.(2025)Chern, Hu, Chern, Kou, Su, Ma, Deng, and Liu]{chern2025thinking}
Ethan Chern, Zhulin Hu, Steffi Chern, Siqi Kou, Jiadi Su, Yan Ma, Zhijie Deng, and Pengfei Liu.
\newblock Thinking with generated images.
\newblock \emph{arXiv preprint arXiv:2505.22525}, 2025.

\bibitem[Comanici et~al.(2025)Comanici, Bieber, Schaekermann, Pasupat, Sachdeva, Dhillon, Blistein, Ram, Zhang, Rosen, et~al.]{Gemini-2.5-flash}
Gheorghe Comanici, Eric Bieber, Mike Schaekermann, Ice Pasupat, Noveen Sachdeva, Inderjit Dhillon, Marcel Blistein, Ori Ram, Dan Zhang, Evan Rosen, et~al.
\newblock Gemini 2.5: Pushing the frontier with advanced reasoning, multimodality, long context, and next generation agentic capabilities.
\newblock \emph{arXiv preprint arXiv:2507.06261}, 2025.

\bibitem[Corbi{\`e}re et~al.(2025)Corbi{\`e}re, Roburin, Montariol, Bosselut, and Alahi]{corbiere2025retrieval}
Charles Corbi{\`e}re, Simon Roburin, Syrielle Montariol, Antoine Bosselut, and Alexandre Alahi.
\newblock Retrieval-based interleaved visual chain-of-thought in real-world driving scenarios.
\newblock \emph{arXiv preprint arXiv:2501.04671}, 2025.

\bibitem[Deepmind(2024)]{Gemini-2.0}
Google Deepmind.
\newblock Introducing gemini 2.0: our new ai model for the agentic era.
\newblock \url{https://blog.google/technology/google-deepmind/google-gemini-ai-update-december-2024/}, 2024.

\bibitem[Deepmind(2025)]{Gemini-2.5-pro}
Google Deepmind.
\newblock Introducing openai o3 and o4-mini.
\newblock \url{https://deepmind.google/models/gemini/pro/}, 2025.

\bibitem[Deng et~al.(2025)Deng, Zhu, Li, Gou, Li, Wang, Zhong, Yu, Nie, Song, et~al.]{deng2025emerging}
Chaorui Deng, Deyao Zhu, Kunchang Li, Chenhui Gou, Feng Li, Zeyu Wang, Shu Zhong, Weihao Yu, Xiaonan Nie, Ziang Song, et~al.
\newblock Emerging properties in unified multimodal pretraining.
\newblock \emph{arXiv preprint arXiv:2505.14683}, 2025.

\bibitem[Duan et~al.(2025)Duan, Fang, Wang, Wang, Huang, Zeng, Li, and Liu]{duan2025got}
Chengqi Duan, Rongyao Fang, Yuqing Wang, Kun Wang, Linjiang Huang, Xingyu Zeng, Hongsheng Li, and Xihui Liu.
\newblock Got-r1: Unleashing reasoning capability of mllm for visual generation with reinforcement learning.
\newblock \emph{arXiv preprint arXiv:2505.17022}, 2025.

\bibitem[Fang et~al.(2025)Fang, Duan, Wang, Huang, Li, Yan, Tian, Zeng, Zhao, Dai, et~al.]{fang2025got}
Rongyao Fang, Chengqi Duan, Kun Wang, Linjiang Huang, Hao Li, Shilin Yan, Hao Tian, Xingyu Zeng, Rui Zhao, Jifeng Dai, et~al.
\newblock Got: Unleashing reasoning capability of multimodal large language model for visual generation and editing.
\newblock \emph{arXiv preprint arXiv:2503.10639}, 2025.

\bibitem[Gao et~al.(2023{\natexlab{a}})Gao, Pi, Zhang, Ye, Zhong, Wang, Hong, Han, Xu, Li, et~al.]{gao2023g}
Jiahui Gao, Renjie Pi, Jipeng Zhang, Jiacheng Ye, Wanjun Zhong, Yufei Wang, Lanqing Hong, Jianhua Han, Hang Xu, Zhenguo Li, et~al.
\newblock G-llava: Solving geometric problem with multi-modal large language model.
\newblock \emph{arXiv preprint arXiv:2312.11370}, 2023{\natexlab{a}}.

\bibitem[Gao et~al.(2023{\natexlab{b}})Gao, Madaan, Zhou, Alon, Liu, Yang, Callan, and Neubig]{gao2023pal}
Luyu Gao, Aman Madaan, Shuyan Zhou, Uri Alon, Pengfei Liu, Yiming Yang, Jamie Callan, and Graham Neubig.
\newblock Pal: Program-aided language models.
\newblock In \emph{International Conference on Machine Learning}, pp.\  10764--10799. PMLR, 2023{\natexlab{b}}.

\bibitem[Google(2025)]{nano-banana}
Google.
\newblock Introducing gemini 2.5 flash image, our state-of-the-art image model.
\newblock \url{https://developers.googleblog.com/en/introducing-gemini-2-5-flash-image/}, 2025.

\bibitem[Guo et~al.(2025)Guo, Yang, Zhang, Song, Zhang, Xu, Zhu, Ma, Wang, Bi, et~al.]{guo2025deepseek}
Daya Guo, Dejian Yang, Haowei Zhang, Junxiao Song, Ruoyu Zhang, Runxin Xu, Qihao Zhu, Shirong Ma, Peiyi Wang, Xiao Bi, et~al.
\newblock Deepseek-r1: Incentivizing reasoning capability in llms via reinforcement learning.
\newblock \emph{arXiv preprint arXiv:2501.12948}, 2025.

\bibitem[Hong et~al.(2025)Hong, Yu, Gu, Wang, Gan, Tang, Cheng, Qi, Ji, Pan, et~al.]{glm}
Wenyi Hong, Wenmeng Yu, Xiaotao Gu, Guo Wang, Guobing Gan, Haomiao Tang, Jiale Cheng, Ji~Qi, Junhui Ji, Lihang Pan, et~al.
\newblock Glm-4.1 v-thinking: Towards versatile multimodal reasoning with scalable reinforcement learning.
\newblock \emph{arXiv e-prints}, pp.\  arXiv--2507, 2025.

\bibitem[Hu et~al.(2024)Hu, Shi, Fu, Roth, Ostendorf, Zettlemoyer, Smith, and Krishna]{hu2024visual}
Yushi Hu, Weijia Shi, Xingyu Fu, Dan Roth, Mari Ostendorf, Luke Zettlemoyer, Noah~A Smith, and Ranjay Krishna.
\newblock Visual sketchpad: Sketching as a visual chain of thought for multimodal language models.
\newblock \emph{Advances in Neural Information Processing Systems}, 37:\penalty0 139348--139379, 2024.

\bibitem[Jiang et~al.(2025)Jiang, Heng, Ye, Yang, Xu, Yan, Zhang, Huang, and Zhang]{jiang2025vlm}
Chaoya Jiang, Yongrui Heng, Wei Ye, Han Yang, Haiyang Xu, Ming Yan, Ji~Zhang, Fei Huang, and Shikun Zhang.
\newblock Vlm-r$^{3}$: Region recognition, reasoning, and refinement for enhanced multimodal chain-of-thought.
\newblock \emph{arXiv preprint arXiv:2505.16192}, 2025.

\bibitem[Li et~al.(2025{\natexlab{a}})Li, Wang, Yue, Cai, Liu, Fu, Guo, Zhu, Sharan, Jia, et~al.]{li2025zebra}
Ang Li, Charles Wang, Kaiyu Yue, Zikui Cai, Ollie Liu, Deqing Fu, Peng Guo, Wang~Bill Zhu, Vatsal Sharan, Robin Jia, et~al.
\newblock Zebra-cot: A dataset for interleaved vision language reasoning.
\newblock \emph{arXiv preprint arXiv:2507.16746}, 2025{\natexlab{a}}.

\bibitem[Li et~al.(2025{\natexlab{b}})Li, Wu, Zhang, Xia, Mao, Dong, Vuli{\'c}, and Wei]{li2025imagine}
Chengzu Li, Wenshan Wu, Huanyu Zhang, Yan Xia, Shaoguang Mao, Li~Dong, Ivan Vuli{\'c}, and Furu Wei.
\newblock Imagine while reasoning in space: Multimodal visualization-of-thought.
\newblock \emph{arXiv preprint arXiv:2501.07542}, 2025{\natexlab{b}}.

\bibitem[Lu et~al.(2023)Lu, Bansal, Xia, Liu, Li, Hajishirzi, Cheng, Chang, Galley, and Gao]{lu2023mathvista}
Pan Lu, Hritik Bansal, Tony Xia, Jiacheng Liu, Chunyuan Li, Hannaneh Hajishirzi, Hao Cheng, Kai-Wei Chang, Michel Galley, and Jianfeng Gao.
\newblock Mathvista: Evaluating mathematical reasoning of foundation models in visual contexts.
\newblock \emph{arXiv preprint arXiv:2310.02255}, 2023.

\bibitem[MathArena(2025)]{HMMT}
MathArena.
\newblock hmmt-feb-2025.
\newblock \url{https://huggingface.co/datasets/MathArena/hmmt_feb_2025}, 2025.

\bibitem[OpenAI(2024)]{GPT-4o}
OpenAI.
\newblock Gpt-4o system card.
\newblock \url{https://openai.com/index/gpt-4o-system-card/}, 2024.

\bibitem[OpenAI(2025{\natexlab{a}})]{GPT-4.1}
OpenAI.
\newblock Introducing gpt-4.1 in the api.
\newblock \url{https://openai.com/index/gpt-4-1/}, 2025{\natexlab{a}}.

\bibitem[OpenAI(2025{\natexlab{b}})]{GPT-o3}
OpenAI.
\newblock Introducing openai o3 and o4-mini.
\newblock \url{https://openai.com/index/introducing-o3-and-o4-mini/}, 2025{\natexlab{b}}.

\bibitem[Pan et~al.(2025)Pan, Wu, Bu, Shen, Li, Wang, Li, Tang, Xiao, Wu, et~al.]{pan2025unlocking}
Kaihang Pan, Yang Wu, Wendong Bu, Kai Shen, Juncheng Li, Yingting Wang, Yunfei Li, Siliang Tang, Jun Xiao, Fei Wu, et~al.
\newblock Unlocking aha moments via reinforcement learning: Advancing collaborative visual comprehension and generation.
\newblock \emph{arXiv preprint arXiv:2506.01480}, 2025.

\bibitem[Seed(2025)]{Seed1.6}
Bytedance Seed.
\newblock Seed1.6 tech introduction.
\newblock \url{https://seed.bytedance.com/en/seed1_6}, 2025.

\bibitem[Shao et~al.(2024)Shao, Qian, Xiao, Song, Zong, Wang, Liu, and Li]{shao2024visual}
Hao Shao, Shengju Qian, Han Xiao, Guanglu Song, Zhuofan Zong, Letian Wang, Yu~Liu, and Hongsheng Li.
\newblock Visual cot: Advancing multi-modal language models with a comprehensive dataset and benchmark for chain-of-thought reasoning.
\newblock \emph{Advances in Neural Information Processing Systems}, 37:\penalty0 8612--8642, 2024.

\bibitem[Shi et~al.(2024)Shi, Hu, Bin, Liu, Yang, Ng, Bing, and Lee]{shi2024math}
Wenhao Shi, Zhiqiang Hu, Yi~Bin, Junhua Liu, Yang Yang, See-Kiong Ng, Lidong Bing, and Roy Ka-Wei Lee.
\newblock Math-llava: Bootstrapping mathematical reasoning for multimodal large language models.
\newblock \emph{arXiv preprint arXiv:2406.17294}, 2024.

\bibitem[Sur{\'\i}s et~al.(2023)Sur{\'\i}s, Menon, and Vondrick]{suris2023vipergpt}
D{\'\i}dac Sur{\'\i}s, Sachit Menon, and Carl Vondrick.
\newblock Vipergpt: Visual inference via python execution for reasoning.
\newblock In \emph{Proceedings of the IEEE/CVF international conference on computer vision}, pp.\  11888--11898, 2023.

\bibitem[Team et~al.(2025)Team, Kamath, Ferret, Pathak, Vieillard, Merhej, Perrin, Matejovicova, Ram{\'e}, Rivi{\`e}re, et~al.]{team2025gemma}
Gemma Team, Aishwarya Kamath, Johan Ferret, Shreya Pathak, Nino Vieillard, Ramona Merhej, Sarah Perrin, Tatiana Matejovicova, Alexandre Ram{\'e}, Morgane Rivi{\`e}re, et~al.
\newblock Gemma 3 technical report.
\newblock \emph{arXiv preprint arXiv:2503.19786}, 2025.

\bibitem[vals.ai(2025{\natexlab{a}})]{AIME}
vals.ai.
\newblock Aime benchmark.
\newblock \url{https://www.vals.ai/benchmarks/aime-2025-09-08}, 2025{\natexlab{a}}.

\bibitem[vals.ai(2025{\natexlab{b}})]{math_500}
vals.ai.
\newblock Math 500 benchmark.
\newblock \url{https://www.vals.ai/benchmarks/math500-08-26-2025}, 2025{\natexlab{b}}.

\bibitem[Wang et~al.(2023)Wang, Ren, Zhou, Lu, Luo, Shi, Zhang, Song, Zhan, and Li]{wang2023mathcoder}
Ke~Wang, Houxing Ren, Aojun Zhou, Zimu Lu, Sichun Luo, Weikang Shi, Renrui Zhang, Linqi Song, Mingjie Zhan, and Hongsheng Li.
\newblock Mathcoder: Seamless code integration in llms for enhanced mathematical reasoning.
\newblock \emph{arXiv preprint arXiv:2310.03731}, 2023.

\bibitem[Wang et~al.(2024)Wang, Pan, Shi, Lu, Ren, Zhou, Zhan, and Li]{wang2024measuring}
Ke~Wang, Junting Pan, Weikang Shi, Zimu Lu, Houxing Ren, Aojun Zhou, Mingjie Zhan, and Hongsheng Li.
\newblock Measuring multimodal mathematical reasoning with math-vision dataset.
\newblock \emph{Advances in Neural Information Processing Systems}, 37:\penalty0 95095--95169, 2024.

\bibitem[Wang et~al.(2025{\natexlab{a}})Wang, Pan, Wei, Zhou, Shi, Lu, Xiao, Yang, Ren, Zhan, et~al.]{wang2025mathcoder}
Ke~Wang, Junting Pan, Linda Wei, Aojun Zhou, Weikang Shi, Zimu Lu, Han Xiao, Yunqiao Yang, Houxing Ren, Mingjie Zhan, et~al.
\newblock Mathcoder-vl: Bridging vision and code for enhanced multimodal mathematical reasoning.
\newblock \emph{arXiv preprint arXiv:2505.10557}, 2025{\natexlab{a}}.

\bibitem[Wang et~al.(2025{\natexlab{b}})Wang, Li, Yin, Ran, and Liu]{wang2025mv}
Peijie Wang, Zhong-Zhi Li, Fei Yin, Dekang Ran, and Cheng-Lin Liu.
\newblock Mv-math: Evaluating multimodal math reasoning in multi-visual contexts.
\newblock In \emph{Proceedings of the Computer Vision and Pattern Recognition Conference}, pp.\  19541--19551, 2025{\natexlab{b}}.

\bibitem[Wang et~al.(2025{\natexlab{c}})Wang, Gao, Gu, Pu, Cui, Wei, Liu, Jing, Ye, Shao, et~al.]{wang2025internvl3}
Weiyun Wang, Zhangwei Gao, Lixin Gu, Hengjun Pu, Long Cui, Xingguang Wei, Zhaoyang Liu, Linglin Jing, Shenglong Ye, Jie Shao, et~al.
\newblock Internvl3. 5: Advancing open-source multimodal models in versatility, reasoning, and efficiency.
\newblock \emph{arXiv preprint arXiv:2508.18265}, 2025{\natexlab{c}}.

\bibitem[Wang et~al.(2025{\natexlab{d}})Wang, Wang, Cheng, Fei, Ding, Guo, Tao, and Qiu]{wang2025visuothink}
Yikun Wang, Siyin Wang, Qinyuan Cheng, Zhaoye Fei, Liang Ding, Qipeng Guo, Dacheng Tao, and Xipeng Qiu.
\newblock Visuothink: Empowering lvlm reasoning with multimodal tree search.
\newblock \emph{arXiv preprint arXiv:2504.09130}, 2025{\natexlab{d}}.

\bibitem[Wei et~al.(2022)Wei, Wang, Schuurmans, Bosma, Xia, Chi, Le, Zhou, et~al.]{wei2022chain}
Jason Wei, Xuezhi Wang, Dale Schuurmans, Maarten Bosma, Fei Xia, Ed~Chi, Quoc~V Le, Denny Zhou, et~al.
\newblock Chain-of-thought prompting elicits reasoning in large language models.
\newblock \emph{Advances in neural information processing systems}, 35:\penalty0 24824--24837, 2022.

\bibitem[Yang et~al.(2025)Yang, Wen, Ding, Liu, Chu, Song, Rao, Yi, Li, Zang, et~al.]{keye}
Biao Yang, Bin Wen, Boyang Ding, Changyi Liu, Chenglong Chu, Chengru Song, Chongling Rao, Chuan Yi, Da~Li, Dunju Zang, et~al.
\newblock Kwai keye-vl 1.5 technical report.
\newblock \emph{arXiv preprint arXiv:2509.01563}, 2025.

\bibitem[Zhang et~al.(2025)Zhang, Zhong, Xia, Yu, Li, He, Shu, Liu, She, Wang, et~al.]{zhang2025cmmcot}
Guanghao Zhang, Tao Zhong, Yan Xia, Zhelun Yu, Haoyuan Li, Wanggui He, Fangxun Shu, Mushui Liu, Dong She, Yi~Wang, et~al.
\newblock Cmmcot: Enhancing complex multi-image comprehension via multi-modal chain-of-thought and memory augmentation.
\newblock \emph{arXiv preprint arXiv:2503.05255}, 2025.

\bibitem[Zhang et~al.(2024{\natexlab{a}})Zhang, Jiang, Zhang, Lin, Guo, Qiu, Zhou, Lu, Chang, Qiao, et~al.]{zhang2024mathverse}
Renrui Zhang, Dongzhi Jiang, Yichi Zhang, Haokun Lin, Ziyu Guo, Pengshuo Qiu, Aojun Zhou, Pan Lu, Kai-Wei Chang, Yu~Qiao, et~al.
\newblock Mathverse: Does your multi-modal llm truly see the diagrams in visual math problems?
\newblock In \emph{European Conference on Computer Vision}, pp.\  169--186. Springer, 2024{\natexlab{a}}.

\bibitem[Zhang et~al.(2024{\natexlab{b}})Zhang, Wei, Jiang, Guo, Li, Zhang, Tong, Liu, Zhou, Wei, et~al.]{zhang2024mavis}
Renrui Zhang, Xinyu Wei, Dongzhi Jiang, Ziyu Guo, Shicheng Li, Yichi Zhang, Chengzhuo Tong, Jiaming Liu, Aojun Zhou, Bin Wei, et~al.
\newblock Mavis: Mathematical visual instruction tuning with an automatic data engine.
\newblock \emph{arXiv preprint arXiv:2407.08739}, 2024{\natexlab{b}}.

\bibitem[Zhou et~al.(2023)Zhou, Wang, Lu, Shi, Luo, Qin, Lu, Jia, Song, Zhan, et~al.]{zhou2023solve}
Aojun Zhou, Ke~Wang, Zimu Lu, Weikang Shi, Sichun Luo, Zipeng Qin, Shaoqing Lu, Anya Jia, Linqi Song, Mingjie Zhan, et~al.
\newblock Solving challenging math word problems using gpt-4 code interpreter with code-based self-verification.
\newblock \emph{arXiv preprint arXiv:2308.07921}, 2023.

\end{thebibliography}
\bibliographystyle{iclr2026_conference}

\clearpage
\appendix
\section{Dataset}
\subsection{Data Collection}
\begin{wrapfigure}{r}{0.48\textwidth}
    \centering
\vspace{-1.2cm}
\includegraphics[width=0.47\textwidth]{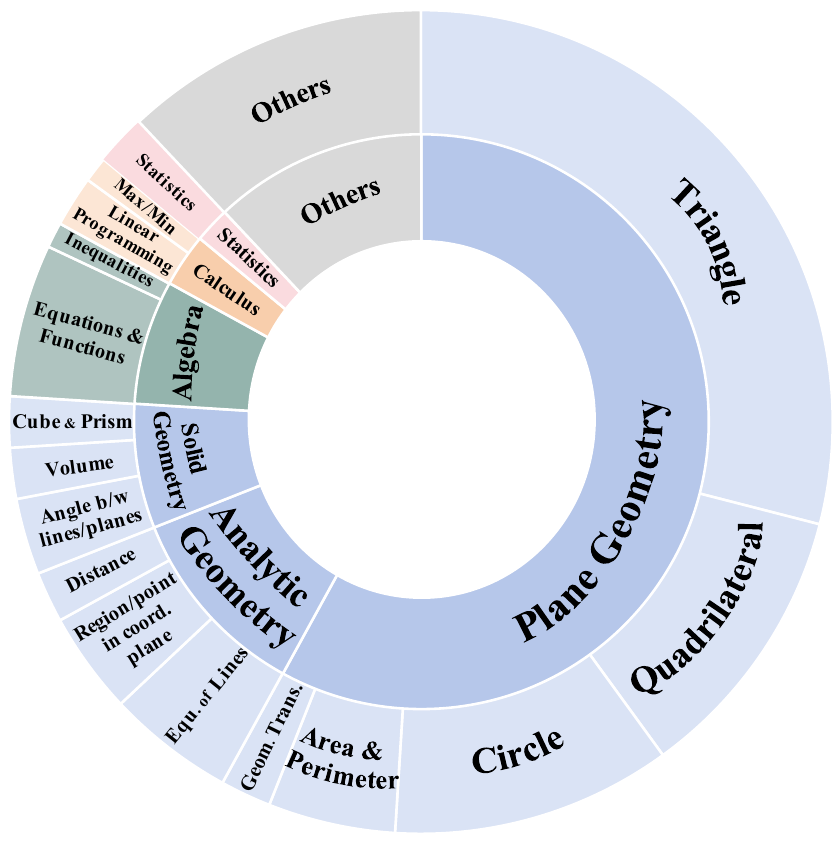} 
\vspace{-0.3cm}
\caption{\textbf{Distribution of Knowledge Types in
Math-VR Dataset.} Geometry constitutes
the majority of problems (76\%), with Algebra
and Calculus comprising 12\%.}
\vspace{-0.3cm}
    \label{fig:dataset_category}
\end{wrapfigure}
We first collect approximately 900k secondary school–level mathematical problems and their corresponding solutions from public websites.
Each sample contains at least one image in the solution explanation. 
These images vary in nature: some consist solely of texts about the problem, others depict mathematical figures such as geometric diagrams or function plots, while a portion are non-informative or irrelevant images. We label these images with Qwen2.5-VL-72B and retain only those questions containing mathematical images and discard the rest.
To improve readability and ensure dataset consistency, we leverage GPT-4.1 to extract text from the remaining purely textual images. Subsequently, we task GPT-4.1 to rewrite each original sample into two standardized markdown files, representing the question and solution separately. In the final step, GPT-4.1 verifies these files to identify and remove incomplete or incoherent samples.
Furthermore, to improve the accessibility of our dataset, we use GPT-4.1 to translate the samples into English, resulting in 178,150 bilingual question-solution pairs.
\begin{wraptable}{r}{0.48\textwidth}
\small
\centering
\renewcommand{\arraystretch}{1.0} 
\vspace{-1cm}
\caption{\textbf{Key Statistics for Math-VR Dataset}}
\vspace{-1mm}
\label{tab:statistics_dataset}
\setlength\tabcolsep{1.6pt} 

\begin{tabular}{lc}
    \toprule
    \textbf{Statistics} & \textbf{Number}  \\
    \midrule
    \textbf{Total Unique Samples} & \textbf{89,075} \\
     \ - Training  & 86,575 \\
     \ - Testing & 2500  \\[-3pt]
    \multicolumn{2}{l}{\ ------------------------------------------------} \\ [-3pt]
     \ - Text questions & 29\% \\
     \ - Multimodal questions & 71\%  \\
     \midrule
     \ - Single-part questions        & \textbf{51\%} \\
   
     \ \ \ \ - Multiple-choice    & 40\% \\
     \ \ \ \ - Answer-based       & 55\% \\
     \ \ \ \ - Proof-based        & 5\% \\
     \ - Multi-part questions         & \textbf{49\%}  \\

     \ \ \ \ - Multiple-choice    & 3\% \\
     \ \ \ \ - Answer-based       & 72\% \\
     \ \ \ \ - Proof-based      & 25\% \\ [-3pt]
    \multicolumn{2}{l}{\ \ \ \ --------------------------------------------} \\ [-3pt]
     \ \ \ \ - Two sub-questions          & 51\% \\
     \ \ \ \ - Three sub-questions        & 40\% \\
     \ \ \ \ - Four sub-questions         & 7\% \\
     \midrule
     \textbf{Question length (text tokens)} \\
      \ - Average & 131.64 \\
     \midrule
     \textbf{Solution length (text tokens)} \\
      \ - Average & 510.76 \\
      \midrule
     \textbf{Multimodal Question Image} \\
      \ - Average number& 1.05 \\
      \ - Average resolution & 208x139 \\
      \midrule
     \textbf{Solution Image} \\
      \ - Average number & 1.15 \\
      \ - Average resolution & 199x151 \\   

    \bottomrule
    \end{tabular}
\vspace{-0.5cm}
\end{wraptable}

\subsection{Dataset Categorization}
In Math-VR dataset, approximately 29\% of the questions are expressed using text only, while the remaining 71\% are demonstrated with both text and mathematical images. Both types of questions require visual mathematical reasoning. 
In addition to organizing our dataset based on the modality of the questions, we can also classify them according to different knowledge types.
First, we randomly sample several thousand question-solution pairs and input them into an VLM (here, GPT-4.1), which generates initial labels including a root knowledge type, a sub-knowledge type and a primary knowledge point, for each pair. Using these labels alongside formal definitions of mathematical concepts, we construct a hierarchical knowledge tree. This taxonomy then serves as a framework to categorize all 90k unique samples in our dataset.
Based on this categorization, the dataset is divided into four main knowledge domains: Geometry, Algebra, Calculus, and Statistics. Geometry constitutes the largest portion, 81\% of all samples. Within Geometry, there are three primary subcategories: Plane Geometry, Solid Geometry, and Analytic Geometry. These subcategories are further broken down into specific knowledge points such as Triangle, Circle, Quadrilateral, and Area \& perimeter calculation.
This hierarchical categorization is visually summarized in the pie chart shown in Figure~\ref{fig:dataset_category}.

\subsection{Dataset Statistics}
We summarize the statistics of our dataset in Table~\ref{tab:statistics_dataset}. The dataset comprises approximately 90k unique questions covering a wide range of types, including both single-part and multipart questions, with two- and three-part questions being the most common. It also includes multiple-choice, answer-based, and proof-based questions, among which answer-based questions are the most prevalent. On average, each solution contains at least one image, reflecting the importance of using images to support the reasoning process.

\section{Benchmark}

\subsection{Templates used in VLM-based Evaluation Pipeline }
We GPT-4.1 as our VLM evaluation tool.
Figure~\ref{tab:temp_meta} presents the template we use to prompt the VLM for generating the meta information of each sample.
Figure~\ref{tab:temp_eval} displays the template used by the VLM to evaluate the model response to each question.

\subsection{Manual Sample Selection and Meta Information Verification}
\begin{figure}[ht]
    \makebox[\linewidth]{
        \includegraphics[width=1.0\linewidth]{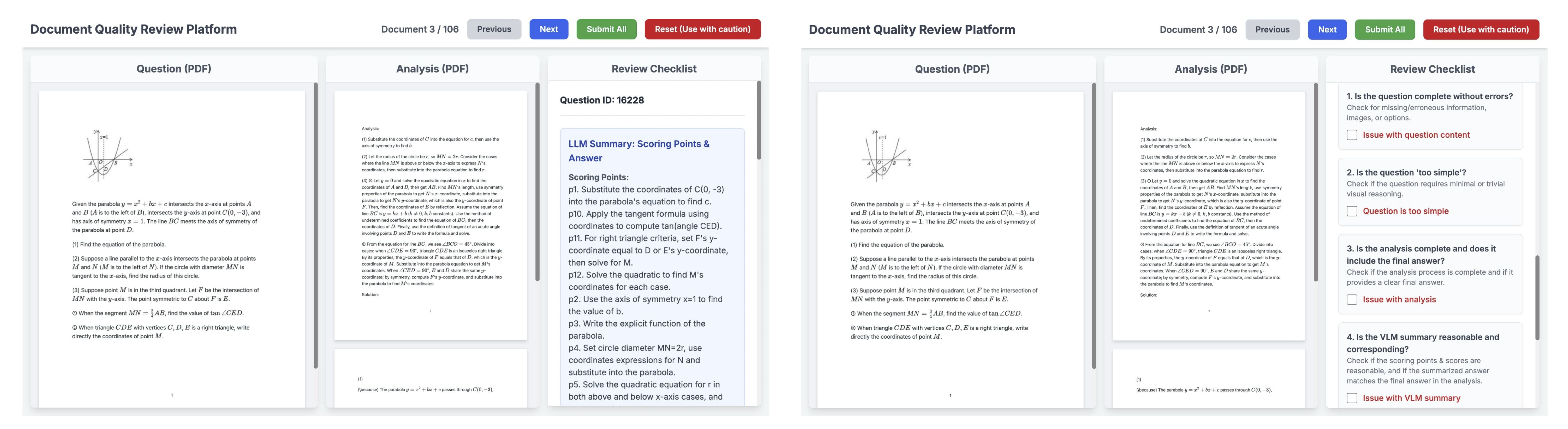}
    }
    \caption{\textbf{The GUI interface for manual sample selection and meta information verification}. 
    This interface displays the Question and Analysis files on the left and center panels, with the Review Checklist on the right. \textit{\textbf{Left}}: The meta information extracted for each sample is displayed on the right panel. \textit{\textbf{Right}}: Scrolling down reveals the items for annotators to check and flag.}
    \label{fig:GUI}
    \vspace{-10pt}
\end{figure}

We combine the selection of samples that require non-trivial visual reasoning with the verification of meta information into a single manual review process. Figure~\ref{fig:GUI} shows the interface used for this review. The interface displays the question, solution, and meta information all within one view.
Annotators are tasked with flagging the unqualified samples. For the remaining samples, they verify that the extracted answers match the final answers in the ground-truth solutions, and that the identified scoring points and their values are reasonable and consistent.
All manual verifications involved in constructing the benchmark are performed by 15 senior college students majoring in STEM disciplines.

\begin{figure}[h!]
\centering
\begin{tcolorbox}[colframe=black, colback=gray!20, arc=5mm, boxrule=0.5mm, width=\textwidth]
\begin{tabular}{p{\linewidth}}
\texttt{<USER>:} I will give you the question, analysis, and answer of a mathematical problem with the ID: \{\textcolor{cyan}{idd}\}. 
The problem may consist of a single question or multiple sub-questions. 

1. Summarize the Answers:
Clearly summarize all answers for the entire problem.
Indicate how many answers there are and specify which answer corresponds to which question or sub-question.

2. List Scoring Points:
For each question or sub-question, identify the scoring points.
Scoring points are the critical steps needed to solve the problem, such as applying theorems, making necessary reasoning steps, or performing correct calculations.
Please assign a score (e.g., 1, 2) for each scoring point based on its difficulty 

Present your answer in the following JSON format:
\{"id": [question id], "scoring\_points":[here should be a dictionary with scoring points: 'p1': [scoring point 1], 'p2': [scoring point 2],…… ], 
"scores":[here should be a dictionary with scores for the scoring points: 's1': [score for p1], 's2': [score for p2],…… ], 
"total\_answer": [the number of all answers], "answer\_summary":[here should be a dictionary, the keys should match the questions, the values should be the answers]\}

\textcolor{cyan}{...Question...}

\\
\midrule
\\
\texttt{<ASSISTANT>:} 

\{
    "id": 25006,
    
    "scoring\_points": \{
    
      "p1": "Recognize that AB is the diameter, thus angles ACB and ADB are right angles by the inscribed angle theorem.",
      
      "p2": "Apply the Pythagorean theorem to calculate BC in triangle ACB and BD in triangle ADB.",
      
      "p3": "Determine that triangle ABC is an isosceles right triangle to find angle BAC.",
      
      "p4": "Determine angle BAD in triangle ABD using calculated lengths.",
      
      "p5": "Classify the different configurations (AC and AD on the same or opposite sides of AB), and compute corresponding values for angle CAD.",
      
      "p6": "Combine stepwise results to conclude that the possible values for angle CAD are 15° or 75°."
    \},
    
    "scores": \{
      "s1": 1,
      "s2": 2,
      "s3": 1,
      "s4": 1,
      "s5": 2,
      "s6": 1
    \},
    
    "total\_answer": 1,
    
    "answer\_summary": \{
      "1": "The possible values of angle CAD are 15° or 75°."
    \},
    
    "max\_score": 8
  \}

\end{tabular}
\end{tcolorbox}
\caption{
\textbf{Template used by GPT-4.1 to generate meta information for evaluation.} The text highlighted in \textcolor{cyan}{cyan} should be replaced with information of the specific question.
}
\label{tab:temp_meta}
\end{figure}
\begin{figure}[h!]
\centering
\begin{tcolorbox}[colframe=black, colback=gray!20, arc=5mm, boxrule=0.5mm, width=\textwidth]
\begin{tabular}{p{\linewidth}}
\texttt{<USER>:} 
You are an expert math teacher and grader. Your task is to evaluate a student's solution to a mathematical question and provide a score. 
You will be provided with the mathematical question (which may include multiple sub-questions), its ID, the student's solution, the correct answer, and the maximum possible score for the question below:

\textcolor{cyan}{...Question...}

\{
'id':\{\textcolor{cyan}{question\_id}\}, 
'student\_solution':\{\textcolor{orange}{model\_response}\}, 

'correct\_answer':\{\textcolor{cyan}{correct\_answer}\}, 
'max\_score':\{\textcolor{cyan}{max\_score}\}
\}.

Please follow these steps precisely:

1. Initial Check for Correctness:

 - Thoroughly review the question and the student\_solution to identify the student’s final answer.
 
 - Compare this final answer directly with the provided correct\_answer. 
 
 - If the answers match exactly, award the full max\_score.

2. Partial Credit Evaluation:

 - If the student's answer is not fully correct, evaluate its work for partial credit using the grading rubric: \{'scoring\_points':\{\textcolor{cyan}{scoring\_points}\}, 'point\_values':\{\textcolor{cyan}{point\_values}\}\}.
 
 - Go through each scoring\_point, indicate if the student successfully completed that step.
 
 - Write down all the point\_ids that the student earned and calculate the total score by summing the values of those points.

3. Provide your evaluation in a strict JSON format:

\{
  "id": "string",
  
  "question\_solution\_analysis": "string"
  
  "is\_fully\_correct": "boolean",
  
  "check\_scoring\_point":"string",
  
  "awarded\_points": ["all" OR a list of earned point\_ids like "p1", "p2"],
  
  "final\_score": "number"
\}

Field Explanations:

 - "id": question id.
 
 - "question\_solution\_analysis": Analyze the question requirements and compare the student's answer against the correct\_answer."
 
 - "is\_fully\_correct": True if the student's solution is fully correct, otherwise False.
 
 - "check\_scoring\_point": If fully correct, provide an empty string "". If not fully correct, explain where in the student\_solution each scoring point is fully met or not met.
 
 - "awarded\_points": If fully correct, this should be ["all"]. If partially correct, provide a list containing the fully met point\_ids (e.g., ["p1", "p3"]). If no points met, provide an empty list [].
 
- "final\_score": the max\_score if fully correct, or the sum of partial scores otherwise.

\end{tabular}
\end{tcolorbox}
\caption{
\textbf{Template used by GPT-4.1 to evaluate model response.} The text highlighted in \textcolor{cyan}{cyan} should be replaced with information of the specific question. The orange \textcolor{orange}{model\_response} should be replaced with the response generated by the model given the question. 
}
\label{tab:temp_eval}
\end{figure}

\vspace{-3pt}
\section{Training Details}
We leverage Qwen2.5VL-32B-Instruct as the base model for both MatPlotCode and the CodePlot-CoT model. During training, we pad all input images into squares before resizing them randomly with lengths between 224 and 560. For rendered images in reasoning chain, we pad and embed them at a fixed resolution of 448×448. The training for MatPlotCode is a two-stage process: we first align the visual components by training only the vision encoder (ViT) and the MLP projector for one epoch, and then perform full-parameter fine-tuning for an additional two epochs. We use a batch size of 512 and a learning rate of 2e-5. For the main CodePlot-CoT model, we initialize its weights from our vision-aligned converter after the completion of Stage 1. We then fine-tune this model on our curated SFT dataset for 5000 steps, using a batch size of 256 and a learning rate of 3e-5. All settings are conducted on 32 NVIDIA H200 GPUs, with each training stage taking approximately 36 hours.

\vspace{-3pt}
\section{CodePlot-CoT Samples on Math-VR Benchmark}
\begin{figure}[t!]
\centering
\begin{tcolorbox}[colframe=black, colback=white!20, arc=5mm, boxrule=0.5mm, width=\textwidth]
\begin{tabular}{p{\linewidth}}
\textbf{Model Generated Example:}

\vspace{0.5em}
\textbf{Question:}

\includegraphics[width=0.25\textwidth]{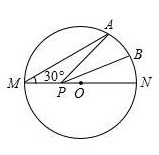}

As shown, $MN$ is the diameter of circle $\bigodot O$ and the radius of $\bigodot O$ is $2$.
Point $A$ lies on $\bigodot O$ and $\angle AMN = 30^{\circ}$.
$B$ is the midpoint of arc $AN$.
$P$ is a movable point on the diameter $MN$.
Find the minimum value of $PA + PB$:\ \underline{\hspace{2.5cm}}.

\vspace{1em}

\textbf{Analysis:}

Reflect point $B$ across line $MN$ to get point $B'$, and connect $AB'$.
By the property of axial symmetry, $AB'$ gives the minimum value of $PA + PB$.
From the properties of central and inscribed angles subtending the same arc:

\begin{itemize}
\item $\angle AON = 2 \angle AMN = 2 \times 30^{\circ} = 60^{\circ}$,
\item By symmetry, $\angle B'ON = \angle BON = 30^{\circ}$.
\end{itemize}

With these, calculate the measure of $\angle AOB'$, then use the Pythagorean theorem to obtain the answer.

\vspace{1em}

\textbf{Solution:}

\includegraphics[width=0.3\textwidth]{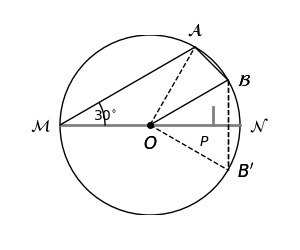}

Reflect point $B$ across $MN$ to get $B'$, and connect $OA$, $OB$, $OB'$, and $AB'$.
The intersection of $AB'$ with $MN$ is the location where $PA + PB$ attains its minimum, thus $PA + PB$ minimum $= AB'$.
\[
\angle AON = 2\angle AMN = 2 \times 30^{\circ} = 60^{\circ}
\]

$B$ is the midpoint of arc $AN$:
\[
\angle BON = \frac{1}{2} \times 60^{\circ} = 30^{\circ}
\]

By symmetry:
\[
\angle B'ON = \angle BON = 30^{\circ}
\]
\[
\angle AOB' = \angle AON + \angle B'ON = 60^{\circ} + 30^{\circ} = 90^{\circ}
\]

Thus $\triangle AOB'$ is an isosceles right triangle:
\[
AB' = \sqrt{2} \cdot OA = 2\sqrt{2}
\]

\textbf{Answer:}\quad $2\sqrt{2}$

\end{tabular}
\end{tcolorbox}
\caption{A multimodal math question form Math-VR benchmark and CodePlot-CoT generated analysis.}
\label{fig:ours_mm}
\end{figure}

\begin{figure}[h!]
\centering
\begin{tcolorbox}[colframe=black, colback=white!20, arc=5mm, boxrule=0.5mm, width=\textwidth]
\begin{tabular}{p{\linewidth}}

\textbf{Model Generated Example:}

\vspace{0.5em}
\textbf{Question:}

\vspace{0.5em}

For an isosceles triangle $\triangle ABC$ with a perimeter of $36\,\text{cm}$, if the altitude to its base $BC$ is $12\,\text{cm}$, what is the value of $\cos B$?

\vspace{0.5em}
A. $\dfrac{1}{2}$ B. $\dfrac{3}{2}$ C. $\dfrac{12}{13}$ D. $\dfrac{5}{13}$

\vspace{1em}

\textbf{Problem Analysis:}

\vspace{0.5em}

Let $AB=x\,\text{cm}$, then $AC=AB=x\,\text{cm}$, and $BC=(36-2x)\,\text{cm}$.
Calculate $BD=DC=\dfrac{1}{2}BC=(18-x)\,\text{cm}$.
In right triangle $ABD$, by the Pythagorean theorem, we have:
\[
x^2=12^2+(18-x)^2.
\]
Solving gives $x=13$, that is, $AB=13\,\text{cm}$, $BD=5\,\text{cm}$.

\vspace{0.5em}
\textbf{Problem Explanation}

\includegraphics[width=0.3\textwidth]{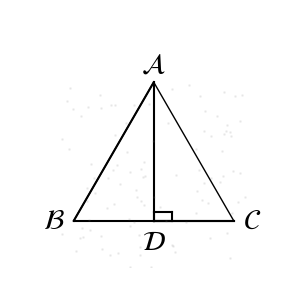}

Let $AB=x\,\text{cm}$, then $AC=AB=x\,\text{cm}$, and $BC=(36-2x)\,\text{cm}$.
Since $AB=AC$ and $AD$ is the altitude, so $BD=DC=\dfrac{1}{2}BC=(18-x)\,\text{cm}$.
In right triangle $ABD$, by the Pythagorean theorem:
\[
AB^2=AD^2+BD^2.
\]
That is,
\[
x^2=12^2+(18-x)^2.
\]
Solving gives $x=13$.
So $AB=13\,\text{cm}$, $BD=5\,\text{cm}$.
Thus $\cos B=\dfrac{BD}{AB}=\dfrac{5}{13}$.

Therefore, the answer is D.

\end{tabular}
\end{tcolorbox}
\caption{A pure textual math question form Math-VR benchmark and CodePlot-CoT generated analysis.}
\label{fig:ours_text}
\end{figure}

We present some samples of CodePlot-CoT on Math-VR Benchmark in Figure~\ref{fig:ours_mm} and~\ref{fig:ours_text}. Figure~\ref{fig:ours_mm} includes a math figure in input, while Figure~\ref{fig:ours_text} is purely text-based inputs. However, both problems require visual reasoning. As shown, our model demonstrates strong reasoning capabilities in both textual and visual reasoning, highlighting our code-driven visual reasoning paradigm.

\vspace{-4pt}
\section{Limitations}
\begin{figure}[h!]
\centering
\begin{tcolorbox}[colframe=black, colback=white!20, arc=5mm, boxrule=0.5mm, width=\textwidth]
\begin{tabular}{p{\linewidth}}
\textbf{Model Generated Example:}

\vspace{0.5em}
\textbf{Question:}

\vspace{0.5em}

If the diagonals of a quadrilateral are equal, then the quadrilateral formed by sequentially connecting the midpoints of its sides must be \underline{\hspace{2.5cm}}.  

\vspace{0.5em}
Options:\quad A.\ Trapezoid \quad B.\ Rectangle \quad C.\ Rhombus \quad D.\ Square  

\vspace{0.5em}

\textbf{Analysis:}  

\vspace{0.5em}
Given that the diagonals of the quadrilateral are equal, by the Midline Theorem of triangles, it follows that all four sides of the resulting quadrilateral are equal, so the quadrilateral formed is a rhombus.  

\vspace{1em}

\textbf{Solution:}  

\includegraphics[width=0.3\textwidth]{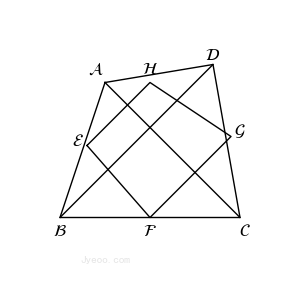}

As shown in the figure, $AC = BD$, and $E$, $F$, $G$, $H$ are the midpoints of segments $AB$, $BC$, $CD$, and $AD$, respectively.  

Since $EH$ and $FG$ are the midlines of $\triangle ABD$ and $\triangle BCD$, respectively, and $EF$ and $HG$ are the midlines of $\triangle ACD$ and $\triangle ABC$, respectively,  

\[
EH = FG = \frac{1}{2}BD, \quad EF = HG = \frac{1}{2}AC
\]

Since $AC = BD$  

\[
EH = FG = EF = HG
\]

Therefore, quadrilateral $EFGH$ is a rhombus. Thus, the answer is C.  

\vspace{0.5em}
\textbf{Answer:}\quad C.\ Rhombus  

\end{tabular}
\end{tcolorbox}
\caption{A pure textual math question form Math-VR benchmark and CodePlot-CoT generated solution. The textual reasoning is correct, but the model generates a slightly imperfect image in reasoning by placing H not on edge AD.}
\label{fig:limit}
\end{figure}

Due to the limitations of data scale and model size, our MatPlotCode has not yet achieved a 100\% fidelity rate on images-to-code conversion, which means that some visual reasoning images are not entirely accurate.
As a result, our final model may also produce slightly imperfect visual reasoning in certain problems.
For instance, in Figure~\ref{fig:limit}, point H should lie on edge AD, but the generated image shows a slight displacement.
We hope that future work will further advance the task of image-to-code conversion, thereby enabling a more robust realization of our paradigm.


\begin{figure}[h!]
\centering
\begin{tcolorbox}[colframe=black, colback=white!20, arc=5mm, boxrule=0.5mm, width=\textwidth]
\begin{tabular}{p{\linewidth}}
\textbf{Math-VR Example:}

\vspace{0.5em}
\textbf{Question:}

\vspace{0.5em}
As shown, $\angle MON = 20^\circ$, A is a point on ray $OM$ with $OA = 4$, D is a point on ray $ON$ with $OD = 8$, C is any point on ray $AM$, and B is any point on segment $OD$. Find the minimum value of the length $AB + BC + CD$ of the broken line $ABCD$.

\includegraphics[width=0.3\textwidth]{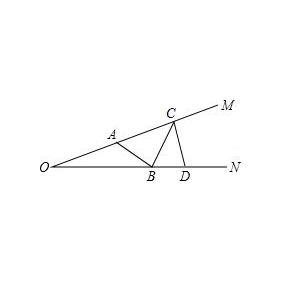}

\textbf{Analysis:}

\vspace{0.5em}
Construct the symmetric points $A'$ and $D'$ of $A$ and $D$ with respect to $ON$ and $OM$, respectively, and connect $A'B$, $CD'$, $A'D'$, $OD'$, and $OA'$.

\includegraphics[width=0.3\textwidth]{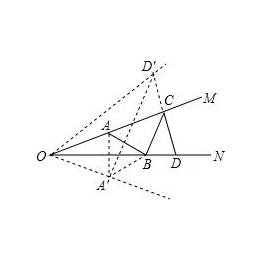}

Then $A'B = AB$, $CD' = CD$, so
\[
AB + BC + CD \geq A'B + BC + CD'.
\]
Clearly,
\[
A'B + BC + CD' \geq A'D'.
\]

Since $\angle A'ON = \angle NOM = \angle MOD' = 20^\circ$, we have $\angle D'OA' = 60^\circ$.

Also, $OA' = OA = 4$, $OD' = OD = 8$, thus
\[
\frac{OA'}{OD'} = \frac{1}{2}.
\]
And since $\cos 60^\circ = \frac{1}{2}$, we have
\[
\cos 60^\circ = \frac{OA'}{OD'},
\]
which shows that $\triangle D'OA'$ is a right triangle, with $\angle OA'D' = 90^\circ$.

Hence,
\[
A'D' = \sqrt{OD'^2 - OA'^2} = \sqrt{8^2 - 4^2} = \sqrt{64 - 16} = \sqrt{48} = 4\sqrt{3}.
\]

\textbf{Answer:} $4\sqrt{3}$.

\end{tabular}
\end{tcolorbox}
\caption{Multimodal input sample from Math-VR.}
\label{fig:dataset_mm}
\end{figure}

\begin{figure}[t!]
\centering
\begin{tcolorbox}[colframe=black, colback=white!20, arc=5mm, boxrule=0.5mm, width=\textwidth]
\begin{tabular}{p{\linewidth}}
\textbf{Math-VR Example:}

\vspace{0.5em}
\textbf{Question:}

\vspace{0.5em}
Let the plane vectors $\overrightarrow{a} = (\cos \alpha, \sin \alpha)$, $\overrightarrow{b} = (\cos(\alpha+\beta), \sin(\alpha+\beta))$, $\overrightarrow{c} = (\cos(\alpha+2\beta), \sin(\alpha+2\beta))$, where $0^\circ < \beta < 180^\circ$. Then ()

\vspace{0.5em}

\textbf{Options:}
\begin{itemize}
    \item[A.] $|\overrightarrow{a} - \overrightarrow{b}| = |\overrightarrow{b} - \overrightarrow{c}|$
    \item[B.] $(\overrightarrow{a} + \overrightarrow{c}) \parallel \overrightarrow{b}$
    \item[C.] If $|\overrightarrow{a} + \overrightarrow{c}| = |\overrightarrow{b}|$, then $\beta = 30^\circ$
    \item[D.] If $\overrightarrow{a} + \overrightarrow{b} + \overrightarrow{c} = \overrightarrow{0}$, then $\beta = 120^\circ$
\end{itemize}

\vspace{0.5em}
\textbf{Analysis:}

\vspace{0.5em}
Because $|\overrightarrow{a}| = |\overrightarrow{b}| = |\overrightarrow{c}| = 1$, we plot $\overrightarrow{OA} = \overrightarrow{a}$, $\overrightarrow{OB} = \overrightarrow{b}$, $\overrightarrow{OC} = \overrightarrow{c}$ respectively on the unit circle.

\includegraphics[width=0.3\textwidth]{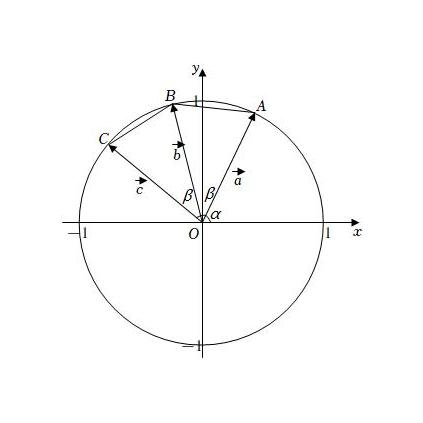}

\textbf{(A)} \quad $|\overrightarrow{a} - \overrightarrow{b}| = |AB|$, \quad $|\overrightarrow{b} - \overrightarrow{c}| = |BC|$.  
Since $\angle AOB = \angle BOC = \beta$, we have $|AB| = |BC|$, i.e.  
$|\overrightarrow{a} - \overrightarrow{b}| = |\overrightarrow{b} - \overrightarrow{c}|.$
Therefore, \textbf{A is correct.}

\vspace{0.5em}
\textbf{(B)} \quad Because $\angle AOB = \angle BOC = \beta$, $OB$ is the angle bisector of $OA$ and $OC$, and $OA = OC = 1$.  
By the vector addition rule, $(\overrightarrow{a} + \overrightarrow{c}) \parallel \overrightarrow{b}$.  
Therefore, \textbf{B is correct.}

\vspace{0.5em}
\textbf{(C)} \quad When $\beta = 60^\circ$, $\triangle OAB$ and $\triangle BOC$ are both equilateral triangles.  
By the parallelogram rule, $\overrightarrow{a} + \overrightarrow{c} = \overrightarrow{b}$, hence
$|\overrightarrow{a} + \overrightarrow{c}| = |\overrightarrow{b}|$.
Thus, $\beta = 60^\circ$, not $30^\circ$, so \textbf{C is incorrect.}

\vspace{0.5em}
\textbf{(D)} \quad Let $(\overrightarrow{a} + \overrightarrow{c}) = \lambda \overrightarrow{b}$, $\lambda \in R$.  
If $\overrightarrow{a} + \overrightarrow{b} + \overrightarrow{c} = \overrightarrow{0}$, then $(\lambda + 1)\overrightarrow{b} = 0$, giving $\lambda = -1$.  
At this point, by the vector rule, $\triangle ABC$ is equilateral, so $\beta = 120^\circ$.  
Hence, \textbf{D is correct.}

\includegraphics[width=0.3\textwidth]{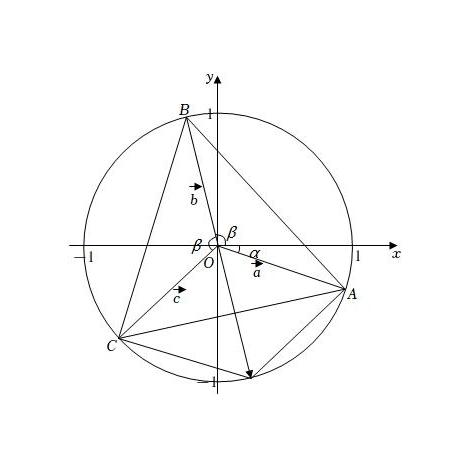}

\textbf{Therefore, the correct options are: A, B, D.}

\end{tabular}
\end{tcolorbox}
\caption{Text input example from Math-VR.}
\label{fig:dataset_text}
\end{figure}


\end{document}